\documentclass[10pt,journal,compsoc]{IEEEtran}
\ifCLASSOPTIONcompsoc
  \usepackage[nocompress]{cite}
\else
  \usepackage{cite}
\fi
\ifCLASSINFOpdf
\else
\fi
\usepackage{multirow}
\usepackage{pifont}
\usepackage{amssymb}
\usepackage{booktabs}
\usepackage{enumerate}
\usepackage{graphicx}
\usepackage{amsmath}
\usepackage{bm}
\usepackage{makecell}
\usepackage{arydshln}
\usepackage{color}
\usepackage[font=small,labelfont=bf,tableposition=top]{caption}
\usepackage[misc]{ifsym}

\DeclareCaptionLabelFormat{andtable}{#1~#2  \&  \tablename~3}
\makeatletter
\newcommand{\otherlabel}[2]{\protected@edef\@currentlabel{#2}\label{#1}}
\makeatother

\usepackage{algorithmic}

\usepackage[linesnumbered,ruled,vlined]{algorithm2e}

\usepackage{subfigure}

\usepackage{float}
\newlength{\bfspace}       
\begin{document}
%
\title{Spatial-Temporal Search for \\ Spiking Neural Networks}
%
%
%
%

\author{
        Kaiwei Che,
        Zhaokun Zhou,
        Li Yuan,
        Jianguo Zhang, 

        Yonghong Tian$^{*}$,~\IEEEmembership{Fellow,~IEEE,}
        and Luziwei Leng$^{*}$
\IEEEcompsocitemizethanks{
\IEEEcompsocthanksitem Kaiwei Che, Zhaokun Zhou, Li Yuan and Yonghong Tian are with Peking University, School of Electronic and Computer Engineering, Shenzhen Graduate School, China, and also with PengCheng Laboratory. E-mail: \{chekaiwei,zhouzhaokun \}@stu.pku.edu.cn, and \{yhtian, yuanli-ece\}@pku.edu.cn.
\IEEEcompsocthanksitem Jianguo Zhang is with Southern University of Science and Technology, China. E-mail: {fwei}@pku.edu.cn.
\IEEEcompsocthanksitem Luziwei Leng is with ACS Lab, Huawei Technologies, Shenzhen, China. E-mail: {lengluziwei}@huawei.com.
\IEEEcompsocthanksitem $^{*}$ Yonghong Tian and Luziwei Leng are corresponding authors.
}
\thanks{Manuscript received November 11, 2023.}}

\IEEEtitleabstractindextext{
\begin{abstract}
Spiking Neural Networks (SNNs) are considered as a potential candidate for the next generation of artificial intelligence with appealing characteristics such as sparse computation and inherent temporal dynamics. By adopting architectures of deep Artificial Neural Networks (ANNs), SNNs achieve competitive performances on benchmark tasks such as image classification. 
However, successful architectures of ANNs are not necessarily optimal for SNNs. In this work, we leverage Neural Architecture Search (NAS) to find suitable architectures for SNNs. Previous NAS methods for SNNs focus primarily on the spatial dimension, with a notable lack of consideration for the temporal dynamics that are of critical importance for SNNs.
Drawing inspiration from the heterogeneity of biological neural networks, we propose a differentiable approach to optimize SNN on both spatial and temporal dimensions.
At spatial level, we have developed a spike-based differentiable hierarchical search (SpikeDHS) framework, where spike-based operation is optimized on both the cell and the layer level under computational constraints.
During the training of SNN, a suboptimal surrogate gradient (SG) function could lead to poor approximations of true gradients, causing the network to enter certain local minima.
To address this problem, we further propose a differentiable surrogate gradient search (DGS) method to evolve local SG functions independently during training. 
At temporal level, we explore an optimal configuration of diverse temporal dynamics on different types of spiking neurons by evolving their time constants, based on which we further develop hybrid networks combining SNN and ANN, balancing both accuracy and efficiency. 
These methods are integrated and verified on a variety of tasks on both static and neuromorphic datasets including image classification and dense prediction.
Our methods achieve comparable classification performance of CIFAR10/100 and ImageNet with accuracies of 96.43\%, 78.96\%, and 70.21\%, respectively.
On event-based deep stereo, our methods find optimal layer variation and surpass the accuracy of specially designed ANNs with 26$\times$ lower computational cost ($6.7\mathrm{mJ}$), demonstrating the potential of SNN in processing highly sparse and dynamic signals.
\end{abstract}

\begin{IEEEkeywords}
Spiking Neural Network, Neural Architecture Search, Surrogate Gradient Optimization, Temporal Dynamics Optimization
\end{IEEEkeywords}}

\maketitle

%
\IEEEdisplaynontitleabstractindextext

%
\IEEEpeerreviewmaketitle


\section{Introduction}\label{sec:introduction}
\IEEEPARstart{I}{nspired} from biological neural networks, spiking neural networks (SNNs) \cite{maass1997networks} have been considered as a potential candidate for the next generation of artificial intelligence, with appealing characteristics such as asynchronous computation, sparse activation, and inherent temporal dynamics. 
However, training deep SNNs is challenging due to the binary spike, which is incompatible with gradient-based backpropagation. To address this challenge, various surrogate gradient (SG) methods have been proposed \cite{bohte2000spikeprop, wu2018spatio, neftci2019surrogate}, where soft relaxed functions are used to approximate the original discontinuous gradient. 
Based on these methods, SNNs have achieved high performance on benchmark image classification tasks such as CIFAR and ImageNet \cite{shrestha2018slayer, wu2019direct, zhang2020temporal, rathi2021diet, zheng2021going}. However, the accuracy of SNN often drops when directly adopting Artificial Neural Network (ANN) architectures, such as ResNet \cite{he2016deep} and VGG networks \cite{simonyan2014very}. With recent improved SNN training methods \cite{li2021differentiable, fang2021deep, deng2022temporal}, the performance gap is decreasing, but remains. This gap is more pronounced for tasks where the network architecture requires more variation, such as dense image prediction \cite{zhu2022event, hagenaars2021self, kim2022beyond}. Directly inheriting sophisticated ANN architectures may not be ideal for SNN, and there is a lack of studies of optimal architectures for spiking neurons given a particular task. 
Topology in the cortex may provide some inspiration.  
Neuroscience has studied macrostructures such as columnar organization \cite{mountcastle1997columnar} and microdynamics such as the interaction between lateral and feedback connections \cite{liang2017interactions}.
However, at the network level, the relationship between connections of neural circuits and cognitive functionalities remains largely unknown. 
Recently, the paradigm for network architecture design in ANN is shifting from human expert driven to almost fully automated \cite{elsken2019neural, liu2021survey}. 
An emerging body of research related to such automated architecture design is called Neural Architecture Search (NAS) \cite{zoph2018learning} and has already surpassed the best manually designed architectures on a variety of tasks \cite{zoph2016neural, zoph2018learning, liu2019auto}. However, the direct application of NAS to SNN is non-trivial considering the unique properties of SNN such as spike-based computation and information processing in both space and time. 

\begin{figure*}[t!]
    \centering
    \includegraphics[width=0.85\linewidth]{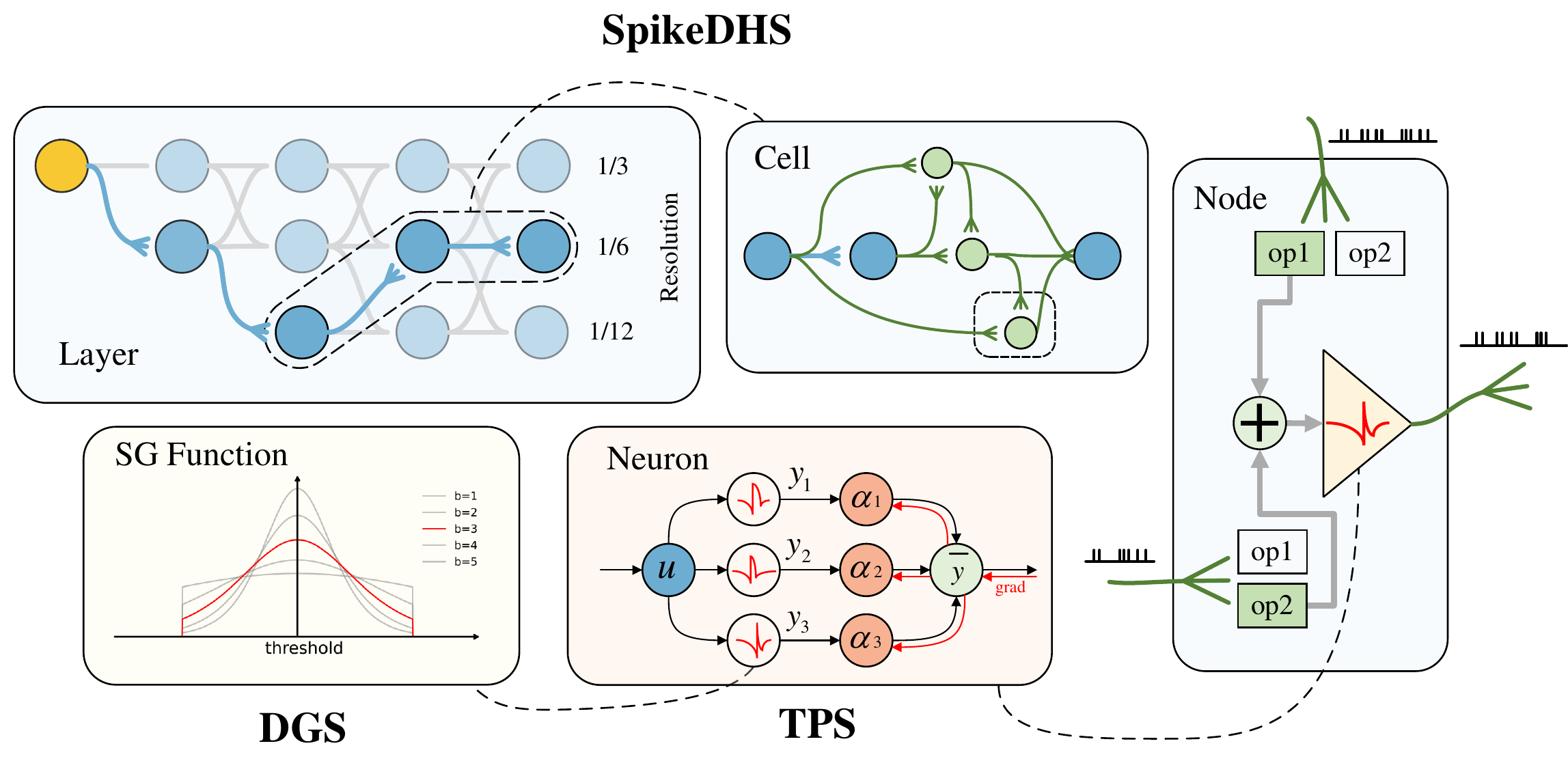}
  \caption{Spatio-temporal search of SNNs, including SpikeDHS, DGS, and TPS. 
  SpikeDHS performs a hierarchical search for SNNs at both the cell and layer level, with the latter optimizing the resolution of the feature map. 
  A cell contains several nodes (e.g. three, shown here as green circles) whose connections are either within the cell or from previous cells, forming a directed acyclic graph. 
  After receiving all operations (e.g., two here), a node is activated by a spiking neuron. 
  Then, DGS explores the optimal surrogate gradient function for spiking neurons during the training process. 
  Based on the searched architecture, TPS explores the optimal temporal dynamics at the neuron level. In practice, the three methods can be combined or used independently.} 
  \label{fig:overall}
  \end{figure*}

Specifically, three questions are raised when constructing a novel SNN for specific tasks:
\begin{itemize}
\item[1.] What is the optimal architecture and set of operations for the specific task?
\item[2.] What shape of surrogate function is optimal for the training procedure?
\item[3.] What is the optimal temporal dynamics for the task?
\end{itemize}

To answer the first question, we establish an architectural foundation and make decisions regarding fundamental operations, such as convolution or max-pooling. 
In particular, we develop a spike-based differentiable hierarchical search (SpikeDHS) framework to find optimal task-specific network architectures under limited computational cost. 
Leveraging the idea of differentiable search \cite{liudarts}, we redesign the search space and information flow according to the principle of spike-based computation. 
For both the cell and the layer search space, we study how to realize this principle towards a fully spiking network. 
To answer the second question, we propose a Differentiable SG Search (DGS) method to explore optimal SG functions, where local SG functions can evolve incrementally and separately during the training phase.
Inspired by the diversity of neural dynamics \cite{hasselmo2006mechanisms,shankar2012scale} and neuron types in the cortex \cite{kandel2000principles,gerstner2014neuronal}, we realize that using fixed temporal parameters and identical neuron across the network might be limiting. 
Therefore, to explore the third question, we propose a Temporal Parameter Search (TPS) method to efficiently find optimal network dynamics, based on which we further develop hybrid networks that combine the advantages of both SNN and ANN. The overall framework is illustrated in Fig. \ref{fig:overall}.

In summary, our contributions are following:
\begin{itemize}
  \item We develop SpikeDHS, a differentiable hierarchical search framework for SNNs, realizing spike-based computation at both the cell and layer level spaces, based on which optimal SNN architectures can be found with limited computational cost.
  \item To improve the gradient approximation of deep SNNs, we propose DGS, a differentiable SG search method to efficiently optimize SG functions locally, which is easy to scale and also effective for binary networks.
  \item We propose a TPS method to efficiently find optimal time constants of neurons, based on which we further develop hybrid networks of ANN and SNN.
  \item Extensive experiments show that our methods outperform SNNs based on sophisticated ANN architectures on image classification of CIFAR10, CIFAR100, and ImageNet datasets. On the event-based deep stereo task, to the best of our knowledge we show the first time that SNN surpasses specially designed ANNs on the Multi Vehicle Stereo Event Camera (MVSEC) \cite{zhu2018multivehicle} in terms of accuracy, network firing rate, and computational cost, demonstrating its advantage in processing highly sparse and dynamic signals with extremely low power and latency.
\end{itemize}

\section{Related Work}

\subsection{Training of SNNs}
The success of deep ANNs in solving machine learning benchmarks has motivated efforts to make SNNs realize similar capabilities, either based on bio-inspired mechanisms \cite{petrovici2016stochastic, neftci2014event, leng2018spiking, mozafari2018first, leng2020solving, korcsak2022cortical} or approximating ANN learning algorithms \cite{bohte2000spikeprop, wu2018spatio, neftci2019surrogate, bellec2020solution}.
At present, two approaches have demonstrated their efficiency, showing the ability to solve hard problems at similar levels as their artificial counterparts, namely ANN-to-SNN conversion \cite{rueckauer2017conversion, bu2021optimal, li2021free, wu2021progressive} and direct training of SNNs with SG \cite{shrestha2018slayer, wu2019direct, zhang2020temporal, li2021differentiable, fang2021deep, deng2022temporal}.
The theoretical soundness of the SG approach for training binary activation networks has been investigated and justified \cite{bengio2013estimating, yin2019understanding}. 
Experiments show that the training of SNNs is robust to the shape of the SG function as long as it satisfies certain criteria, such as the overall scale \cite{zenke2021remarkable}. \textit{Hagenaars et al.}~\cite{hagenaars2021self} shows that an appropriate SG function is critical when the SNN goes deeper. 
\textit{Li et al.}~\cite{li2021differentiable} further improved the performance of the SNN by optimizing the width (or temperature) of a continuous SG function through the guidance of an approximate gradient measured with Finite Difference Gradient (FDG). 
FDG hyperparameters are empirical, and their layer-by-layer nature makes FDG computationally expensive.
Recent work \cite{lian2023learnable} proposes to optimize the shape of the SG function by analyzing the corresponding membrane potential distribution.
Specifically, this work introduces a differentiable SG search (DGS) method to more efficiently explore optimal SG functions during training.

\subsection{Architecture Search of SNNs}
Designing high performance network architectures for specific tasks often requires expert experience and trial-and-error experiments. 
NAS aims to automate this manual process and has recently achieved highly competitive performance in tasks such as image classification \cite{zoph2016neural, zoph2018learning, liu2018progressive, real2019regularized}, object detection \cite{zoph2018learning, chen2019detnas, guo2020hit}, and semantic segmentation \cite{liu2019auto, nekrasov2019fast, lin2020graph}, etc. 
However, searching over a discrete set of candidate architectures often results in a massive number of potential combinations, leading to explosive computational costs. 
The Differentiable Architecture Search (DARTS) \cite{liudarts} and its variants \cite{xu2019pc, chen2019progressive, chu2020darts} address this problem by using a continuous relaxation of the search space, which enables learning a set of architecture coefficients by gradient descent, and has achieved state-of-the-art performance using orders of magnitude fewer computational resources \cite{liudarts, liu2019auto, cheng2020hierarchical}. 
The success of NAS methods in ANN has inspired the development of automated architecture optimization methods \cite{shen2024evolutionary}. 
AutoSNN \cite{na2022autosnn} studied pooling operations for downsampling in SNNs and applied NAS to reduce the total number of spikes. 
SNASNet\cite{kim2022neural} applied NAS to improve SNN initialization and explore backward connections. 
However, both works only searched for different SNN cells or combinations of them under a fixed network backbone, and their application is limited to image classification. 
The recurrent SNN \cite{wang2023evolving} is evolved as a weight-agnostic neural network \cite{gaier2019weight} and applied it to robot locomotion tasks. 
\textit{Shen et al.}~\cite{shen2023brain} extended the search space to excitatory and inhibitory LIF neurons and employed spike-timing-dependent plasticity (STDP) to tune the network architecture. The very recent work of \cite{xie2023efficient} used a genetic algorithm to search an SNN with mixed spiking neurons under a VGG architecture to improve its performance in classification tasks. 
\textit{Yan et al.}~\cite{yan2024sampling} reduces computation by randomly sampling spatial and temporal topologies, avoiding gradient-based optimization, but lacks theoretical guarantees despite low variance in results.
\textit{Yan et al.}~\cite{yan2024efficient} improves the accuracy-cost tradeoff by using a single-path NAS in a branchless spiking supernet, greatly reducing computation and search time.

\subsection{Temporal Dynamics and Hybrid Networks of SNNs}
Traditional SNNs usually adopted fixed temporal hyperparameters, which limited the temporal diversity of neurons. To overcome this limitation, researchers have introduced parametric spiking neurons \cite{zimmer2019technical,wu2021liaf,fang2021incorporating,rathi2021diet}. These methods parameterize partial or whole parameters of a neuron using direct gradient optimization. No prior research has explored search methods for optimizing SNN temporal dynamics. This work introduces a Temporal Parameter Search (TPS) method to efficiently identify optimal network dynamics.
Several recent works \cite{yang2019dashnet,lee2020spike,10347028} have explored the combination of ANN and SNN, constructing hybrid networks by exploiting the advantages of both sides, i.e. high precision computation based on floating-point activation and low computational cost induced by sparse spiking activation.
However, these approaches relied on handcrafted architecture design, which may not result in optimal integration of ANN and SNN. 
We expand the search domain of our TPS framework to include both dynamic spiking neurons and static artificial neurons with floating-point precision, allowing for optimal hybrid architecture design.

\subsection{Event-based Task with SNNs}
\label{sec:event_task}
Inspired by biological retina, event camera \cite{gallego2020event} captures instantaneous changes of pixel intensity at microsecond resolution. 
Compared to traditional frame-based cameras, it covers a higher dynamic range (120dB) and provides a low-power solution for vision tasks in high-speed scenarios. 
Further combination with neuromorphic processors \cite{merolla2014million, roy2019towards, 9695196, 9767613, 10019594} can create a fully neuromorphic system, realizing extremely low power and low latency sensing. 
However, learning from highly sparse and asynchronous events is challenging. 
Given its inherently asynchronous dynamics, SNN is an ideal candidate for such a task, and recently a number of works have applied it to event-based problems such as classification \cite{kugele2020efficient, li2021free, li2024efficient}, tracking \cite{yang2019dashnet}, detection \cite{kim2020spiking,zhang2024automotive}, semantic segmentation \cite{kim2022beyond, zhang2024accurate}, optical flow estimation \cite{lee2020spike, hagenaars2021self, 8660483}, and reconstruction \cite{9695196, 10019594, 10130595}, etc. 
Multi-view event-based deep stereo solves the problem of 3D scene reconstruction based on pixel differences of the same physical point from event streams obtained from multiple views. 
Given the complexity of the problem, it has been addressed by specially designed deep ANNs. 
Several works use additional information such as camera motion to generate sparse depth maps \cite{zhu2018realtime, zhou2018semi}. 
Estimating dense disparity images from sparse event inputs is more challenging. 
A recent work \cite{tulyakov2019learning} addresses this problem by using an event queue method that encodes events into event images through 3D convolution, followed by an hourglass network to estimate disparity. Based on \cite{tulyakov2019learning}, \cite{ahmed2021deep} further improves the local contours of the estimated disparity using image reconstruction. 
\textit{Mostafavi et al.}~\cite{mostafavi2021event} creates feature pyramids with multi-scale correlation learned from a cycle of grayscale images and event inputs. 
Inspired by biological neuron models, \textit{Zhang et al.}~\cite{zhang2022discrete} develops discrete time convolution to encode events with temporal dynamic feature maps. \textit{Ran{\c{c}}on et al.}~\cite{ranccon2022stereospike} applies SNN to this problem with a handcrafted U-net structure. However, its performance is suboptimal compared to ANN models using geometric volumes.

\section{Spiking Neurons and Training of SNN}
In this section, we introduce several spiking neurons and the direct training method of SNN that will be used by the spatial-temporal search framework in the next section. The neuron are introduced in the order of integrate-and-fire (IF) neuron, leaky integrate-and-fire (LIF) neuron, and adaptive threshold integrate-and-fire (ALIF) neuron \cite{bellec2018long,bellec2020solution}, with increasing complexity in leaky behavior and threshold dynamics. For the direct training of the SNN, we adopt the widely used Spatial-Temporal Backpropagation (STBP) \cite{wu2018spatio}.
\subsection{Spiking Neurons}
\subsubsection{IF Neuron}
The IF neuron is a relatively simple spiking neuron where the 
membrane potential is defined by a summation of the gated past membrane potential and the current input, which can be described by the following iterative equation:
\begin{align}
    u^{t} &= u^{t-1}(1-y^{t-1}) + I^{t}, \\
    y^{t} &= H(u^{t}-V_{\mathrm{th}}),
\label{eq0:IF}
\end{align}
where $t$ denote the time step, $u$ is the membrane potential, and $y$ is the spike output defined by the Heaviside function $H$. $y^{t} = 1$ if $u^{t}$ exceeds the threshold $V_{\mathrm{th}}$, otherwise $y^{t} = 0$. $I$ denotes the synaptic input with $I^{t} = \sum_{j} w_{j} y^{t, n-1}_j$ where $w$ is the weight. The spiking activity of the last time step also serves as a gate for the information of the past membrane potential.

\subsubsection{LIF Neuron}
Based on the IF neuron, the more biologically plausible LIF neuron adds leaky behavior to the membrane potential by partially inheriting the past membrane potential gated by its spiking activity from the last time step. The LIF neuron with hard reset can be described by the following iterative equation:
\begin{align}
    u^{t} &= \tau u^{t-1}(1-y^{t-1}) + I^{t}, \\
    y^{t} &= H(u^{t}-V_{\mathrm{th}}),
\label{eq1:LIF}
\end{align}
where $\tau \in (0,1)$ is the leaky factor of the membrane potential. The other variables are defined as in the IF neuron.
In this study, we set the default values of $\tau = 0.2$ and $V_{th} = 0.5$. The TPS method is applied to automatically optimize the $\tau$.

\subsubsection{ALIF Neuron}
The ALIF neuron takes into account the neuronal adaptive excitatory behavior observed in neocortex \cite{bellec2018long} and implements a self-adaptive threshold on the LIF neuron, which can be described by the following iterative form:
\begin{align}
    u^{t} &= \tau u^{t-1}(1-y^{t-1}) + I^{t},  \\
    y^{t} &= H(u^t - A^t), \\
    A^{t} &= V_{\mathrm{th}}+\beta a^{t}, \\
    a^{t} &= \tau_{a} a^{t-1}+y^{t-1}, 
\label{eq1:LIF}
\end{align}
where $A^t$ is the adjustable threshold at time $t$, $a^t$ is the cumulative threshold increment, which changes according to the spiking history of the neuron, and $\beta$ is a scaling factor, $\tau_a \in (0,1)$ is the decay factor of $a$.
The adjustable threshold $A$ modulates the spiking rate so that the threshold rises with dense input, which in turn inhibits the neuron from firing, and vice versa, resulting in self-adaptive spiking activity. If the neuron is continuously inactive, $a^{t} \rightarrow 0$ as $t \rightarrow \infty$, the lower bound of $A^t$ is equal to the original threshold $V_{\mathrm{th}}$. In the extreme case when the neuron is continuously spiking, with $y^0=1$ and $a^0=0$, $a^{t}$ can then be derived as:
\begin{equation}
a^{t} = \sum_{i=1}^{t} \tau_{a}^{i-1},
\label{eq:eqac}
\end{equation}
and its upper limit can be derived as $t \rightarrow \infty$, which is equal to $\frac{1}{1-\tau_{a}}$. Therefore, the variation range of the adjustable threshold $A$ is $[V_{\mathrm{th}}, V_{\mathrm{th}}+\frac{\beta}{1-\tau_{a}}]$.
In this study, we set the default values of $\tau = 0.2$, $\tau_a = 0.2$, and $V_\mathrm{th} = 0.5$. The TPS method is then applied to optimize $\tau$ and $\tau_a$.

\subsection{Direct Training of SNN}
Following to STBP, given a loss $L$, the weight update of the SNN is expressed as:
\begin{equation}
    \frac{\partial L}{\partial w}= \sum^{T-1}_{t=0}\frac{\partial L}{\partial y^{t}} \frac{\partial y^{t}}{\partial u^{t}} \frac{\partial u^{t}} {\partial I^{t}} \frac {\partial I^{t}} {\partial w},
    \label{eq:stbp}
\end{equation}
where T is the simulation step, $\frac{\partial y^{t}}{\partial u^{t}}$ is the gradient of the spiking function, which is zero everywhere except at $u = V_{th}$. The SG approach \cite{bohte2000spikeprop, wu2018spatio, neftci2019surrogate} uses continuous functions to approximate the real gradients. Experiments demonstrates that the training of SNN is robust to a variety of SG functions of different shapes, such as rectangular \cite{zheng2021going}, triangular \cite{bellec2018long}, Superspike \cite{zenke2018superspike}, ArcTan \cite{fang2021incorporating}, and exponential curves \cite{shrestha2018slayer}. We adopt the Dspike function from \cite{li2021differentiable}, formulated as:

\begin{equation}
\mathrm{Dspike}(x)=
a\cdot \mathrm{tanh}(b(x-c))+d \text{, if } 0\leq x \leq 1,
\end{equation}
which can cover a wide range of smoothness by changing the temperature parameter $b$, with $\mathrm{Dspike}(x)=1$ or $0$ for $x>1$ or $x<0$. We set $c=V_{\mathrm{th}}$ and determine $a$ and $d$ by setting $\mathrm{Dspike}(0)=0$, $\mathrm{Dspike}(1)=1$.

\section{Spatial-Temporal Search for SNN}
In this section, we first introduce search methods in the spatial dimension, including SpikeDHS for architectural search in Sec. \ref{sec:spikedhs} and DGS for SG search in Sec. \ref{sec:dgs}. Then we introduce the temporal parameter search method, TPS, in Sec. \ref{sec:tps}. The overall framework of our method is depicted in Fig.\ref{fig:overall}. These methods can be integrated or used independently in practice.

\begin{algorithm}[!t]
    \SetProcArgSty{texttt}
    \caption{SpikeDHS}
    \label{algorithm_SpikeDHS}
    \KwIn{Training dataset, epoch $E$, warm-up epoch $E_{\mathrm{w}}$, candidate operations $\{o_i\}_N$, cell weighting factors $\{\alpha_{i}\}_N$, layer weighting factors $\{\beta_{i}\}_N$, and their initializing value $\epsilon$}
    Split training dataset into $\textit{A}$ and $\textit{B}$ randomly \;
    Combine $\{o_{i}\}_N$ to form a mixed cell operation with $\{\alpha_{i}\}_N$ initialized to $\epsilon$ \;
    Combine mixed cell operations to form a mixed layer output with $\{\beta_{i}\}_N$ initialized to $\epsilon$ \;    
    \For{each epoch $e$ from 1 to $E$}{
        \For{each iteration}{
            Update network weights $w$ by $\nabla_{w}L(w,\alpha, \beta)$ on $\textit{A}$\;    
            \uIf{$e > E_{\mathrm{w}}$}{
                Update architecture parameters $\alpha$ and $\beta$ by $\nabla_{\alpha,\beta}L(w,\alpha, \beta)$ on $\textit{B}$
            }
        }
    }
    Derive the final architecture based on the learned $\alpha$ and $\beta$ \;
    \noindent \textbf{return} Optimal architecture.
\end{algorithm}

\subsection{Spike-based Differentiable Hierarchical Search}

SpikeDHS can be divided into two parts: cell level search and layer level search. We first introduce the formulation and the spiking activation position of a cell. Next, we explain how these cells are used to construct layers automatically. Fig. \ref{fig:overall} provides a visual representation, and Algorithm \ref{algorithm_SpikeDHS} summarizes the pseudo-code for SpikeDHS.

\label{sec:spikedhs}
\subsubsection{Cell Level Search}
\label{sec:cell_search}
\noindent \textbf{Cell formulation.}
A cell contains multiple nodes whose connections are either within the cell or from previous cells, forming a directed acyclic graph, as shown in Fig. \ref{fig:overall}. In our setting, each node can receive inputs from previous nodes as well as from two previous cells. The cell forms its output by concatenating all the outputs of its nodes. The activation function of the node is implemented by a spiking neuron, denoted as:
\begin{equation}
  x_j = f(\sum_{i < j}o^{(i,j)}(x_i)),
\label{eq:node_sum}
\end{equation}
where $x$ denotes the spiking output of the node and $i,j$ denote node indices (for conciseness, we omit the input from previous cells). $f$ is a spiking neuron taking the sum of all operations as input, 
$o^{(i,j)}$ is the operation associated with the directed edge connecting node $i$ and $j$. 
During the search, each edge is represented by a weighted average of candidate operations, and the information flow connecting node $i$ and node $j$ is defined as:
\begin{equation}
    \bar{o}^{(i,j)}(x) = \sum_{o \in O^{(i,j)}}\frac{\mathrm{exp}(\alpha_o^{(i,j)})}{\sum_{o \in O^{(i,j)}} \mathrm{exp}(\alpha_o^{(i,j)})}o(x), 
\label{eq:alpha_op}
\end{equation}
where $O^{(i,j)}$ denotes the operation space on edge $(i,j)$ and $\alpha_o^{(i,j)}$ is the weighting factor of operation $o$, which is a trainable continuous variable. At the end of search, a discrete architecture is selected by replacing each mixed operation $\bar{o}^{(i,j)}$ with the most likely operation $o^{(i,j)} = \mathop{\mathrm{max}}\limits_{o \in O^{(i,j)}} \alpha_o^{(i,j)}$.

\begin{figure}[t!]
    \centering
    \includegraphics[width=1\linewidth]{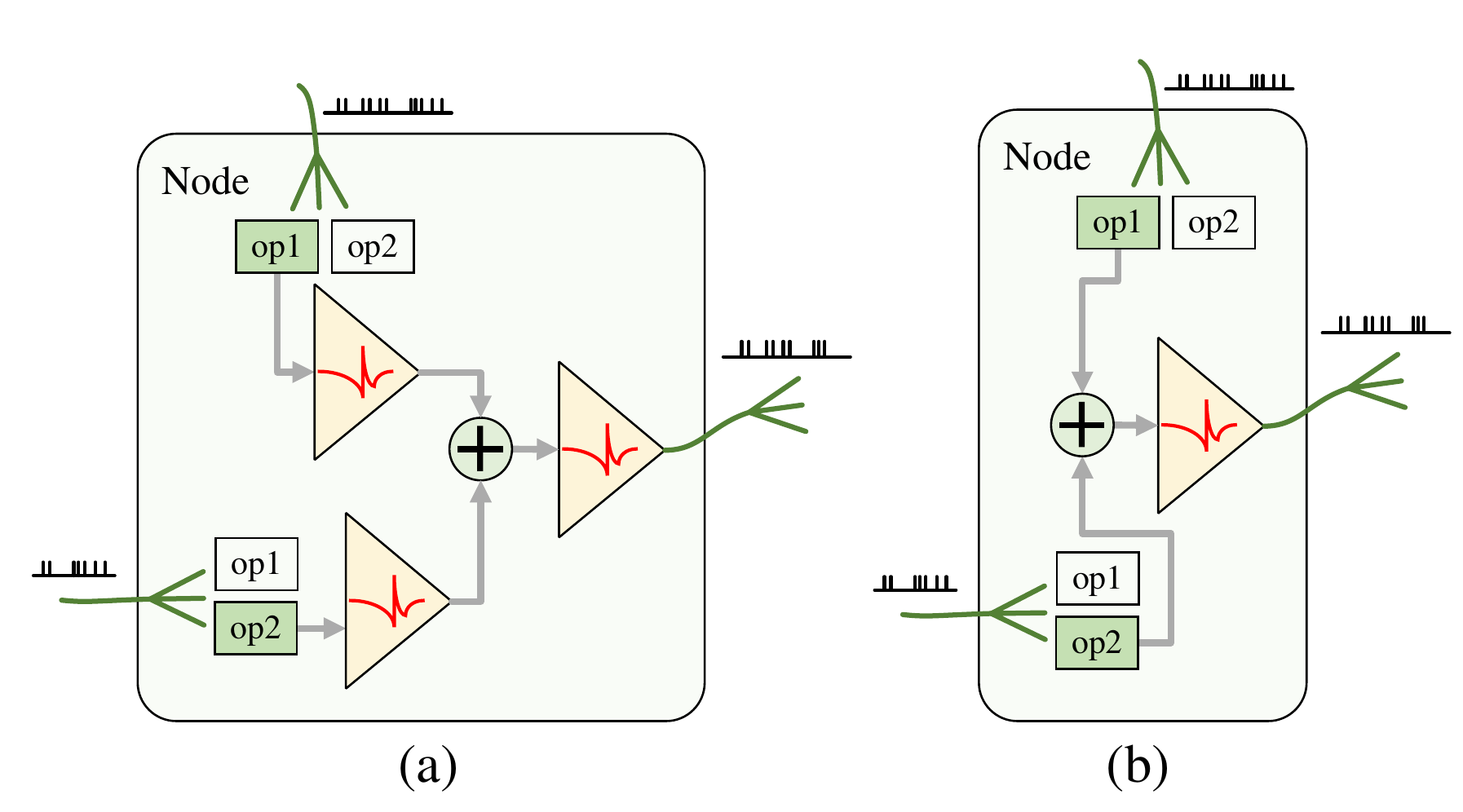}
    \caption{Spiking activation position. (a) Mixed operation at the spike activation; (b) Mixed operation at the membrane potential. Two operations are shown here as an example.}
    \label{fig:mix_mem}
  \end{figure}

\vspace{\bfspace} \noindent \textbf{Spiking activation position.}    
Note that for spiking neurons, unlike artificial neurons, the membrane potential serves as an additional latent variable. Therefore, during search we can mix the operation either at the spike activation or at the membrane potential, as depicted in Fig.\ref{fig:mix_mem}.

Specifically, for mixed operation at the spike activation, Eq. \ref{eq:alpha_op} is replaced with $\bar{y} = \bar{o}$ and $y = o$. The mixed operation then becomes:
\begin{equation}
    \bar{y} = \sum_{k \in K}\frac{e^{\alpha_k}}{Z} y_k, 
\label{eq:alpha_y}
\end{equation}
where $Z = \sum_{k}^{K} e^{\alpha_k}$ is the normalizing constant, $K$ denotes the number of candidate operations on an edge and $k$ is the index of the operation. Given the loss $L$, assuming $T=1$ for simplicity, the gradient of $\alpha_k$ is derived as:
\begin{align}
    \frac{\partial L}{\partial \alpha_k} &= \sum_{t=0}^{T-1} \frac{\partial L}{\partial f^t} \frac{\partial f^t}{\partial \bar{y}^t} \frac{\partial \bar{y}^t}{\partial \alpha_k}  \nonumber \\
    & = \frac{\partial L}{\partial f} \frac{\partial f}{\partial \bar{y}} \frac{e^{\alpha_k}}{Z}\Bigg[ y_k -  \frac{\sum_{k}^{K} (e^{\alpha_k}y_k)}{Z} \Bigg]  \nonumber\\
    & = \frac{\partial L}{\partial f} \frac{\partial f}{\partial \bar{y}} \frac{e^{\alpha_k}}{Z}\Big( y_k -  \bar{y} \Big),
    \label{eq:mix_spk}
\end{align}
where $f$ is the spiking activation of the node from Eq. \ref{eq:node_sum} and for $\frac{\partial f}{\partial \bar{y}}$ we apply a fixed SG function. As Eq. \ref{eq:mix_spk} shows, the gradient of $\alpha_k$ stops at $y_k -  \bar{y}$, so for candidate operations we can use different SG functions for the corresponding spike activation. However, the learning signal of $\alpha$ is also filtered by an additional SG function $\frac{\partial f}{\partial \bar{y}}$ at the node, which could cause additional noise during learning. In addition, when applied for DGS method, the contribution of candidate weights will be filtered by the consecutive spike activation. If the difference between the original weight and candidate weights is minor, they could lead to the same spike activation and in the extreme case $y_{g_i} - \bar{y}$ could be 0 (see Sec. \ref{sec:dgs} for more details).

For mixed operation at the membrane potential, Eq. \ref{eq:alpha_op} is replaced by $\bar{I} = \bar{o}$ and $I = o$. The mixed operation is then expressed as:
\begin{equation}
    \bar{I} = \sum_{k \in K}\frac{e^{\alpha_k}}{Z} I_k.    
\label{eq:alpha_y}
\end{equation}
Assuming $T=1$ for simplification, the gradient of $\alpha_k$ is then derived as:
\begin{align}
    \frac{\partial L}{\partial \alpha_k} &= \sum_{t=0}^{T-1} \frac{\partial L}{\partial y^t} \frac{\partial y^t}{\partial u^t} \frac{\partial u^t}{\partial \bar{I}^t} \frac{\partial \bar{I}^t}{\partial \alpha_k}  \nonumber\\
    &= \frac{\partial L}{\partial y} \cdot g(u) \cdot 1 \cdot \frac{e^{\alpha_k}}{Z} \Big( I_k -  \bar{I}_k \Big).
    \label{eq:mix_mem}
\end{align}
As Eq. \ref{eq:mix_mem} shows, $\frac{\partial y}{\partial u}$ adopts a unified SG function $g(u)$ for different candidate operations which limits the exploration of diverse SG functions. However, since $I_k$ directly depends on the weight, when applied for DGS it strictly reflects different changes of the original weight, and thus transmitting more accurate learning signals for $\alpha_k$. Also, in forward path this leads to a more concise node with one less spiking activation, as shown in Fig. \ref{fig:mix_mem}. We adopt mixed at the the membrane potential in the following experiments.
In previous work \cite{liudarts} a prepossessing step is required on operations between nodes and cells of previous layers to align the dimension of feature maps. We merge this step with subsequent candidate operations to reduce model complexity and improve inference speed. 

\subsubsection{Layer level search}
\noindent \textbf{Layer formulation.}
Task specific knowledge has been proved to be helpful in speeding up the search process and improving network performance. For classification task, we adopt a fixed downsampling structure, with normal cells and reduction cells searched separately.
For dense image prediction where high resolution output is needed and network architecture requires more variation, we implement differentiable search on the layer level as proposed in \cite{liu2019auto}. A set of scalars $\{\beta\}$ are trained to weight different potential layer resolutions and they are updated together with $\alpha$. 
Given a loss function $L$, the update rule of $\{\alpha, \beta\}$ follows a bi-level optimization \cite{liudarts} based on two disjoint training sets \textit{A} and \textit{B} from the training set:
\begin{itemize}
\item Update network weights $\noindent \textbf{w}$ by $\nabla_{\noindent \textbf{w}}L($\noindent \textbf{w}$,\alpha, \beta)$ on $\textit{A}$
\item Update architecture parameters $\alpha$ and $\beta$ by $\nabla_{\alpha,\beta}L($\noindent \textbf{w}$,\alpha, \beta)$ on $\textit{B}$
\end{itemize}

We use first-order approximation to speed up the search process. After search, an optimal structure is decoded from a pre-defined $L$-layer trellis, as shown in Fig. \ref{fig:overall}. We decode the discrete cell structure by retaining two strongest afferent edges for each node. As to network structure, we decode it by finding the maximum probability path between different layers.

To maintain binary feature maps, for upsampling layers, we use nearest interpolation. Batch normalization \cite{ioffe2015batch} is used in the search and retraining phase. 
In test inference, parameters of batch normalization are merged into convolution weights and biases \cite{ioffe2015batch, rueckauer2017conversion}, leading to a full spike-based network.

\subsection{Differentiable Surrogate Gradient Search}
\label{sec:dgs}

\begin{algorithm}[!t]
    \SetProcArgSty{texttt}
      \caption{DGS}
      \label{algorithm_DGS}
      \KwIn{Training dataset, training epoch $E$, training iteration $I$, DGS iteration $I_g$, SG function $g$, candidate SG functions ${\{g_i\}}_N$, SG weighting factors ${\{\alpha_{g_i}\}}_N$ and their initializing value $\epsilon$, epoch interval for DGS $e_D$}
        \For{all $e$ = 1, 2, ..., $E$-th epoch}{
            \For{all $i$ = 1, 2, ..., $I$-iteration}{
                Collect training data and labels, update weights $w$ based on SG function $g$;
            }    
            \uIf{$e/e_{D}=\mathrm{Int.}$}{
                \For{all $j$ =1, 2, ..., $I_g$-iteration}{
                Initialize ${\{\alpha_{g_i}\}}_N$ to $\epsilon$,
                update weights $w$ with $\nabla_{g_i, w}L$ based on ${\{g_i\}}_N$, obtain associated weights ${\{w_{g_i}\}}_N$; 
                
                Combine ${\{w_{g_i}\}}_N$ to form mixed operation, update ${\{\alpha_{g_i}\}}_N$ with $\nabla_{\alpha_{g_i}, {\{w_{g_i}\}}_N}L$. 
                }
                Update $g$ to $\{g_i|i = \mathop{\mathrm{argmax}}_{g_i} \ \langle \alpha_i \rangle \}$
            }
        }
        \noindent \textbf{return} trained network.
     \end{algorithm}

A recent work \cite{hagenaars2021self} shows that a suitable SG function is critical when the SNN goes deeper, and \cite{li2021differentiable} demonstrates that by optimizing the width (or temperature) of the SG function the performance of SNN can be improved. 
Continuous relaxation through gradient descent is an efficient approach to explore diverse operations on the same path, inspired from this idea, we propose DGS to parallelly optimize local SGs for SNN. The psuedo code of the algorithm is summarized in Algorithm. \ref{algorithm_DGS}.

In the retraining phase, with certain epoch intervals,
we associate each operation path with $N$ candidate SG functions, ${\{g_i\}}_N$, based on which we update the weight (or $N$ copies of the weight) separately, leading to ${\{w_{g_i}\}}_N$. 
These weights are then combined to form a mixed operation weighted by a set of factors ${\{\alpha_{g_i}\}}_N$ through a softmax function, described as:
\begin{equation}
    \hat{I} = \sum_{i \in N}\frac{\mathrm{exp}(\alpha_{g_i})}{\sum_{j \in N} \mathrm{exp}(\alpha_{g_j})} I_i \text{ , with    } {I}_i = w_{g_i}x.
\label{eq:diffb}
\end{equation} 
We then update ${\{\alpha_{g_i}\}}_N$ through the loss of the mixed operation output. 
This process is repeated for multiple batches and finally we update the original SG to $\{g_i|i = \mathop{\mathrm{argmax}}_i \ \langle \alpha_i \rangle \}$, with $\langle \cdot \rangle$ denoting the average over batches.
Note that $w_{g_i}$ can be obtained either by repeatedly calculate for each $g_i$, or directly estimating from the gradient of the original SG, $\nabla_{g, w}L$, if ${\{g_i\}}_N$ are linear to $g$. 
The intuition behind DGS is that the updated value of $\alpha_{g_i}$ indicates the contribution of $w_{g_i}$ in decreasing the loss. So $\{g_i|i = \mathop{\mathrm{argmax}}_{g_i} \ \langle \alpha_i \rangle \}$, which leads to the best updated weight, could be the most suitable SG function for the original local weight.
Note that the difference between DGS and the SG search in Sec. \ref{sec:cell_search} is that the former aims to optimize SG function for the local weight, while the latter is essentially a search of different operation types.

\subsection{Temporal Parameter Search}
\label{sec:tps}
Previous NAS methods for SNNs mainly focus on the spatial dimension while neglecting the temporal dynamics, which is critical for SNNs. In particular, the time constants of temporal variables and the reset mechanism together determine the temporal dynamics of general LIF neuron. In this work, we focus on the optimization of time constants.

\begin{algorithm}[!t]
    \SetProcArgSty{texttt}
    \caption{TPS}
    \label{algorithm_TPS}
    \KwIn{Training dataset, maximum iteration $I_{\text{max}}$, iteration step $s$,  candidate temporal parameter $\tau_k \in \{\tau + \Delta \tau, \tau, \tau - \Delta \tau\}$, weighting factors $\{{a}_i\}$, and their initializing value $\epsilon$}
    
    Combine $\{\tau + \Delta \tau, \tau, \tau - \Delta \tau\}$ to form a mixed operation with $\{{a}_i\}_N$ initialized to $\epsilon$ \;
    
        \For{each iteration from 0 to $I_{\text{max}}$}{
            Update network weight $w$ by $\nabla_{w}L(w,a)$\;
            Update $\{a\}_N$ with $\nabla_{a}L(w,a)$\;

            \uIf{every $s$ iteration}{
                Update $\tau$ to $\{\tau_i = \mathop{\mathrm{argmax}}_{\tau_i} \langle {a}_i \rangle\}$\;
                Reset $\{a\}_N$ to $\epsilon$;
            }
        }

    \noindent \textbf{return} Optimal temporal parameter $\tau$\;
\end{algorithm}

\vspace{\bfspace} \noindent \textbf{Formulation.} 
The original LIF dynamics are shown in Eq. \ref{eq1:LIF}. Considering the current layer, the equation can be simplified by ignoring $n$, resulting in:
\begin{align}
    u_k^{t} &= \tau_k u_k^{t-1}(1-y^{t-1}) + I^{t}, \\
    y_k^{t} &= H(u_k^{t}-V_{\mathrm{th}}),
\label{eq:TPS}
\end{align}
where $k$ denotes the specific candidate $\tau_k$ used in this case. We then introduce a control parameter $a$ to combine the different outputs $y_k^{t}$.
\begin{align}
\label{eq:tps_param}
    \bar{y}^t = \sum_{k \in K}\frac{\mathrm{exp}({a}_k)}{\sum_{k \in K} \mathrm{exp}({a}_k)} y_k^t.
\end{align}
The search space of TPS is designed as $\tau_k \in \{\tau + \Delta \tau, \tau, \tau - \Delta \tau\}$. The process is as follows: at each step, TPS optimizes its control parameter $a$ to determine which candidate $\tau$ is optimal, and subsequently replaces the original $\tau$ value with the new optimal $\tau$ value. Following each step, $a$ is reset to its initial value. After certain number of steps, the search process concludes and the optimal time constants $\tau$ is obtained.
The psuedo code of TPS is summarized in Algorithm \ref{algorithm_TPS}.

\begin{figure}[t!]
    \centering
    \includegraphics[width=1\linewidth]{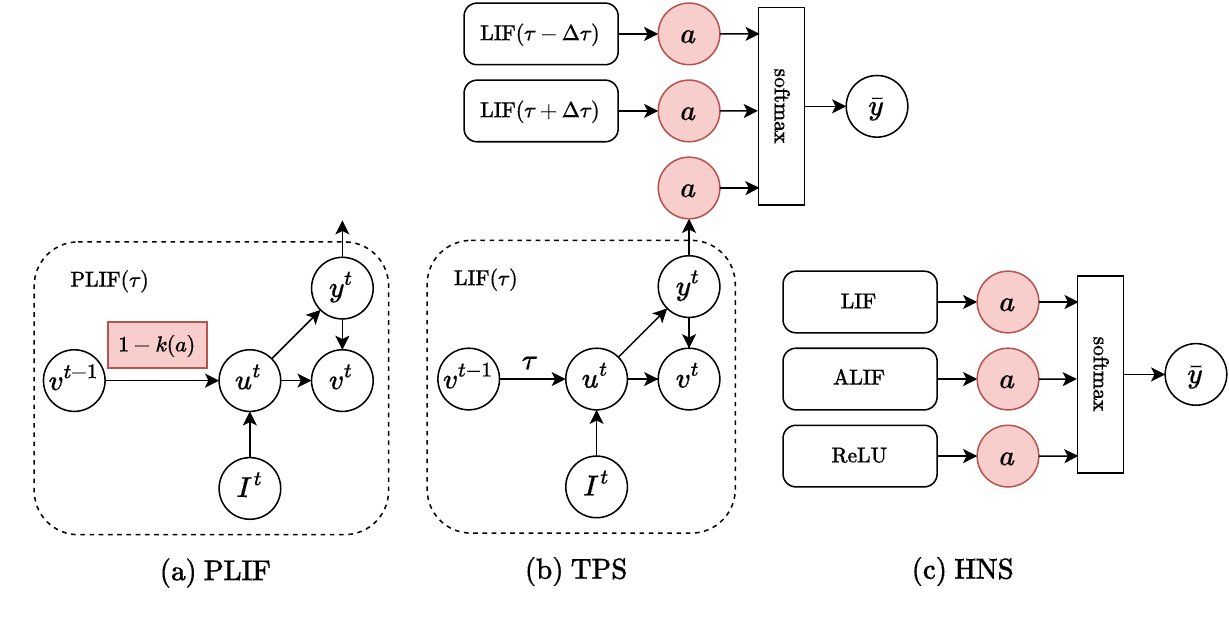}
    \caption{Difference beteen the direct gradient method (a) PLIF and our methods (b) TPS and (c) HNS. PLIF parameterizes temporal parameters through direct gradient optimization. In contrast, TPS combines neurons in a mixed operation, where each neuron's contribution is weighted by a factor ($a$). Over multiple steps, these factors are updated based on gradients, and the neuron with the highest $a$ value is selected as the optimal choice. HNS follows a similar process to TPS but integrates both artificial and spiking neurons.}
    \label{fig:tps}
  \end{figure}

\vspace{\bfspace} \noindent \textbf{Gradient of TPS.}
The dynamics of the LIF model using the TPS method are illustrated in Fig. \ref{fig:tps}. We further analyze the gradient of TPS. Given the loss $L$, the gradient of $a$ can be expressed as:
\begin{align}
    \frac{\partial L}{\partial {a}_k} &= \sum_{t=0}^{T-1} \frac{\partial L}{\partial \bar{y}^t} \frac{\partial \bar{y}^t}{\partial {a}_k},  \\
    \frac{\partial y^t}{\partial {a}_k} &= \frac{e^{{a}_k}}{\sum_k e^{{a}_k}} (y_k^t - \bar{y}^t).
    \label{eq:gradient_tps}
\end{align}

Most previous methods optimize temporal parameters using direct gradient methods. For comparison, we consider the optimization of the PLIF \cite{fang2021incorporating} neuron to illustrate the direct gradient training process. For PLIF neuron, the membrane potential charge process can be expressed:
\begin{equation}
    \label{eq:PLIF}
    u^{t} = (1-k(a))u^{t-1} + I^t,
\end{equation}
where $a$ is parametric membrane parameter, $k(a)=\frac{1}{1+exp(-a)}$. The optimization of $a$ can be described as:
\begin{align}
    \frac{\partial L}{\partial a} &= \sum_{t=0}^{T-1} \frac{\partial L}{\partial y^t} \frac{\partial y^t}{\partial u^t} \frac{\partial u^t}{\partial a}, \\
    \frac{\partial u^t}{\partial a} &= -u^{t-1}k^{\prime}(a) + \frac{\partial u^t}{\partial u^{t-1}} \frac{\partial u^{t-1}}{\partial a}.
\end{align}

\vspace{\bfspace} \noindent \textbf{Analysis.} TPS has the following advantages compared to direct gradient-based training approaches:
 
\vspace{\bfspace} \noindent 1. \textbf{More explicit.} 
As shown in Eq. \ref{eq:gradient_tps}, the gradient of ${a}_k$ depends on the gradient of subsequent layer, $\frac{\partial L}{\partial \bar{y}^t}$, and the difference $y_k^t - \bar{y}^t$. The term $y_k^t - \bar{y}^t$ represents the deviation between the candidate output $y_k^t$, driven by the candidate $\tau_k$, and the desired output $\bar{y}^t$. Using the output difference highlights the direct influence of $\tau$ on the output. Conversely, in the PLIF case, the gradient of $a$ is determined by the gradient of $\frac{\partial u^t}{\partial u^{t-1}} \frac{\partial u^{t-1}}{\partial a}$ and $-u^{t-1}k^{\prime}(a)$, which are more implicit, relying more on the previous state $u^{t-1}$ than on output differences caused by varying $\tau$.

\vspace{\bfspace} \noindent 2. \textbf{Broader range.} 
Owing to its explicit formulation, our method accounts for a broader range of values. If the change in $\tau$ (denotes $\Delta \tau$) is sufficiently small so that the difference between $y_k^t$ and $\bar{y}^t$ is negligible, the gradient of ${a}_k$ will approach zero. A broader step is therefore required to detect differences, indicating that our method considers the influence of $\tau$ over a broader range. In contrast, direct gradient methods tend to focus on small, neighboring values, which can lead to convergence to local optima and increased sensitivity to initial conditions.

\vspace{\bfspace} \noindent 3. \textbf{Less noise.} 
Direct gradient training introduces noise due to the surrogate function. In the PLIF approach, the derivative $\frac{\partial y}{\partial u}$ is approximated by an SG function, which could introduce noise in the learning of $\tau$. In contrast, our method directly calculates $\sum_t^T \frac{\partial L}{\partial \bar{y}^t} \frac{\partial \bar{y}^t}{\partial {a}_k}$, thus avoiding the noise associated with surrogate approximations.

\subsection{Hybrid Network Search} 
\label{sec:hns}
Extending upon TPS, we further propose an ANN-SNN hybrid network search (HNS) framework to improve network spatial representation using artificial non-spiking neurons while preserving sparsity and temporal dynamics of spiking neurons. While direct gradient training methods only optimize networks under fixed neuron type, HNS can optimize a mixed use of artificial neurons and spiking neurons by simply using their activation difference as an explicit indicator, enabling the exploration of efficient algorithms for neuromorphic hardware supporting hybrid networks \cite{2019Towards}.

\vspace{\bfspace}
\noindent \textbf{Formulation.}
In Eq. \ref{eq:tps_param}, $y_k$ represents the output of the neuron with TPS focusing on the differences between various $y_k$. 
Eq. \ref{eq:gradient_tps} highlights how these differences contribute to the gradient. To expand the search space to include ANN-SNN hybrids, we can simply introduce artificial non-spiking neurons into the original spiking neuron search space. In HNS, the search space is defined as $y_k \in \{y_\text{ANN}, y_\text{SNN}\}$, where $y_\text{ANN}$ represent artificial non-spiking neurons (we use the ReLU neuron) and $y_\text{SNN}$ represents the spiking neuron.

\noindent \textbf{Energy loss function.}
Integrating floating-point operations increases energy consumption compared to fully spiking networks. To balance energy consumption and performance in diverse scenarios, we introduce an energy loss function. 
\begin{align}
    \mathcal{L}_{\text{eg}} &= \sum_{l \in L} \sum_{k \in K} \frac{\mathrm{exp}({a}_k)}{\sum_{k} \mathrm{exp}({a}_k)} \rho_k, \\
    \mathcal{L} &=  \mathcal{L}_{\text{ac}} + \lambda \mathcal{L}_{\text{eg}},
  \end{align}
where $\mathcal{L}_{\text{eg}}$ is the energy loss function and $L_{\text{ac}}$ is the original loss function, $\lambda$ controls the trade-off between accuracy and performance, $\rho$ denotes the energy factor of the neuron, $l$ denotes a specific layer and $\sum_{l \in L} \sum_{k \in K}$ denotes the predicted energy consumption across all layers. 
If $\lambda$ is closer to 1, the energy term has a greater influence on the loss; if closer to 0, the influence of energy on the loss diminishes, and accuracy becomes more dominant.

To determine the value of $\rho$ for different neurons, we refer to a study on $45\mathrm{nm}$ CMOS technology \cite{horowitz20141}, widely used in SNN research \cite{li2021differentiable,rathi2021diet,kim2022beyond}. Here, an accumulate (AC) operation consumes $0.9\mathrm{pJ}$, while a multiply-accumulate (MAC) operation consumes $4.6\mathrm{pJ}$. Assuming a total of $A$ operations, with a time step of $T=6$ and a firing rate of $\mathop{fr}=0.2$, the total energy for a LIF neuron is $\mathop{fr} \cdot T \cdot A \cdot 0.9 = 1.08 \cdot A \, \text{mJ}$. For a ReLU neuron, the energy consumption is $4.6 \cdot A \, \text{mJ}$ since $T=1$. Thus, the energy consumption of SNN is about 4.26 times lower than that of ReLU. We set the energy factors of LIF and ReLU to 1 and 4.26, respectively, to facilitate energy prediction and optimization in hybrid networks. Note that energy efficiency evaluation in SNNs is still under research, with alternative methods such as SynOps \cite{merolla2014million} providing alternative evaluations. 

\subsection{Deep Stereo Estimator and Coding Scheme}
For event-based deep stereo task, we specifically apply a sub-pixel estimator \cite{tulyakov2018practical} and a stacking based on time (SBT) encoding \cite{wang2019event} to improve the performance of the network.

\vspace{\bfspace}
\noindent \textbf{Loss function and estimator.}  We use the mean depth error (MDE), one-pixel-accuracy (1PA), median depth error, and mean disparity error as evaluation metrics. The network produces a matching cost tensor $C$ of size $\frac{d_{max}}{2}\times h\times w$, based on which we estimate a disparity map $\hat D$ using a sub-pixel estimator:
\begin{equation}
    \hat D = \sum_d D(d)\mathop{\mathrm{softmin}}\limits_{d:|\hat{d} - d|\textless \delta}(C_{d,y,x}) \text{ , with }    \hat{d} = \mathop{\mathrm{argmin}}\limits_d (C_{d,y,x}),
\end{equation}
where $\delta = 2$ is an estimator support and $D(d) = 2·d$ is a disparity corresponding to index $d$ in the matching cost tensor.
We use a sub-pixel cross entropy loss \cite{tulyakov2018practical} to train the network, which is described by:
\begin{align}
    \mathcal{L}(\Theta) &= \frac{1}{wh}\sum_{y,x}\sum_{d} \mathrm{Laplace}(D(d)|\mu = D^{GT}_{y,x}, b) \cdot \nonumber \\
    &\qquad \mathrm{log}(\mathop{\mathrm{softmin}}\limits_d (C_{d,y,x})),
\end{align}
where $\mathrm{Laplace}(D |\mu = D^{GT}_{y,x}, b)$ is a discretized Laplace distribution with the mean equal to the ground truth disparity $\mu = D^{GT}_{y,x}$ and diversity $b=2$.

\vspace{\bfspace}
\noindent \textbf{Event Encoding.}
Learning from highly sparse raw events is challenging, and a preprocessing step is often required to encode events. For MVSEC, we use stacking based on time (SBT) \cite{wang2019event}, which merges events into temporally adjacent frames. 
During training, we use multiple consecutive stacks as one input, with an equal number of consecutive ground truth disparities as one label.

\section{Experiments}

\subsection{Image Classification}
\subsubsection{Experimental Setup}

\noindent \textbf{Dataset.} We apply SpikeDHS on the CIFAR10 \cite{cifar} and then retrain on the target datasets including CIFAR10, CIFAR100 and ImageNet. The CIFAR10 and CIFAR100 have 50K/10K training/test RGB images with a spatial resolution of $32\times32$. The ImageNet dataset \cite{deng2009imagenet} contains over 1250k training images and 50k test images. 
For ImageNet, we use a larger variant of the searched network with one more stem layer and two more cells. We use standard preprocessing and augmentation for training as in \cite{he2016deep}. The test image is directly centred and cropped to $224\times224$. 
\begin{table}[t!]
    \renewcommand{\arraystretch}{1.5}
    \caption{Search Space for SpikeDHS, TPS and DGS in image classification tasks.}
    \label{tab:cla_search_space}
    \centering
        \begin{tabular}{cccc}\hline
        \noindent \textbf{SpikeDHS} & \noindent \textbf{DGS} &\noindent \textbf{TPS} &\noindent \textbf{HNS} \\\hline

        \makecell[c]{conv3$\times$3 with $g(b=3) $\\conv3$\times$3 with $g(b=5)$\\skip connection} &
        \makecell[c]{$g_1$\\$g_2$\\...} &
        \makecell[c]{$\text{LIF}(\tau)$} &
        \makecell[c]{$\text{LIF}$\\$\text{ReLU}$} 
        \\\bottomrule
    \end{tabular}
\end{table}
\begin{table*}[t!]
    \renewcommand{\arraystretch}{1.3}
    \centering
    \caption{Comparison on CIFAR. ${}^\mathrm{D}$: DGS method. ${}^\mathrm{T}$: TPS method. ${}^\mathrm{H}$: HNS method. NoP: Number of parameters.}
    \label{tab:cifar}
    \begin{tabular}{cccccc}
    \toprule Methods & Architecture & NoP & T & CIFAR10 & CIFAR100 \\
    \midrule
    \cite{zhang2020temporal}TSSL-BP & CIFARNet & - & 5 & $91.41$ & -\\
    \cite{rathi2021diet}Diet-SNN & ResNet-20 & - & 10/6 & $92.54$ & $64.07$ \\
    \cite{zheng2021going}STBP-tdBN & ResNet-19 & 13M &6 & $93.16$ & $71.12^{\pm 0.57}$\\
    \cite{deng2022temporal}TET & ResNet-19 & 13M & 6 & $94.50^{\pm 0.07}$ & $74.72^{\pm 0.28}$ \\
    \cite{li2021differentiable}Dspike & ResNet-18 & 11M & 6 & $94.25^{\pm 0.07}$ & $74.24^{\pm 0.10}$\\
    \cite{kim2022neural}SNASNet & SNASNet-Bw & - & 8 & $94.12^{\pm 0.25}$  & $73.04^{\pm 0.36}$ \\
    \cite{na2022autosnn}AutoSNN & AutoSNN (C=128) & 21M & 8 & $93.15$ & $69.16$ \\
    \cite{wu2021tandem}Tandem Learning & CifarNet & - & 8 & $90.98$ & $-$ \\
    \cite{wu2021progressive}Tandem Learning & VGG-11 & 133M & 16 & $91.24$ & $-$ \\
    \cite{yan2024sampling}STTS & TANet-Tiny & 7M & 4 & $95.10^{\pm 0.09}$ & $76.33^{\pm 0.32}$ \\
    \cite{yan2024efficient}ESNN & ESNN($\lambda=0$) & 23.47M/27.55M & 3 & $94.64$ & $74.78$ \\
    \noindent \textbf{SpikeDHS} & SpikeDHS-CLA & 14M & 6 & \noindent \textbf{95.35$^{\pm 0.05}$} &\noindent \textbf{76.15$^{\pm 0.20}$} \\
    $\noindent \textbf{SpikeDHS}^{\mathrm{D}}$ & SpikeDHS-CLA & 14M & 6 & \noindent \textbf{95.50$^{\pm 0.03}$} & \noindent \textbf{76.25$^{\pm 0.10}$}\\
    $\noindent \textbf{SpikeDHS}^{\mathrm{T}}$ & SpikeDHS-CLA & 14M & 6 & \noindent \textbf{96.36$^{\pm 0.03}$} & \noindent \textbf{77.9$^{\pm 0.15}$}\\ 
    $\noindent \textbf{SpikeDHS}^{\mathrm{H}}$ & SpikeDHS-CLA & 14M & 6 & \noindent \textbf{96.43$^{\pm 0.06}$} & \noindent \textbf{78.96$^{\pm 0.13}$}\\ 
    \bottomrule
\end{tabular}
\end{table*}

\begin{table}[t!]
    \fontsize{7.4pt}{\baselineskip}\selectfont 
    \renewcommand{\arraystretch}{1.3}
    \centering
    \caption{Comparison on ImageNet. ${}^\mathrm{D}$: DGS method. NoP: Number of parameters.}
    \label{tab:imagenet}
    \begin{tabular}{ccccc}
    \toprule  Methods & Architecture & NoP & T & Acc[$\%$] \\
    \midrule
    \cite{wu2019direct}STBP-tdBN & ResNet-34 & 22M & 6 & $63.72$ \\
    \cite{wu2019direct}STBP-tdBN & ResNet-34-large & 86M & 6 & $67.05$ \\
    \cite{rathi2021diet}Diet-SNN & VGG-16 & 138M & 5 & $69.00$ \\
    \cite{wu2021tandem} Tandem Learning & AlexNet & - & 10 & $50.22$\\
    \cite{wu2021progressive} Tandem Learning & VGG-16 & 138M & 16 & $65.08$\\
    \noindent \textbf{SpikeDHS} & SpikeDHS-CLA-large & 58M & 6 & $\noindent \textbf{67.96}$\\
    $\noindent \textbf{SpikeDHS}^{\mathrm{D}}$ & SpikeDHS-CLA-large & 58M & 6 & $\noindent \textbf{68.64}$\\
    $\noindent \textbf{SpikeDHS}^{\mathrm{T}}$ & SpikeDHS-CLA-large & 58M & 6 & $\noindent \textbf{69.50}$\\
    $\noindent \textbf{SpikeDHS}^{\mathrm{H}}$ & SpikeDHS-CLA-large & 58M & 6 & $\noindent \textbf{70.21}$\\
    \bottomrule
\end{tabular}
\end{table}

\vspace{\bfspace}
\noindent \textbf{SpikeDHS.} 
The search space of SpikeDHS includes 4 nodes (n4) within a cell and a limited number of candidate operations to reduce the search time, as shown in Table \ref{tab:cla_search_space}.
For the network architecture, we adopt an 8 layers fixed downsampling architecture proposed in \cite{liudarts}. We use an auxiliary loss as in \cite{liudarts}, with a weight of 0.4. In the search phase, the training set is equally divided into two subsets for bi-level optimisation. For retraining, the standard training/test split is used. The search phase takes $50$ epochs with mini-batch size $50$, the first $15$ epochs are used to warm up the convolution weights. We use SGD optimizer with a momentum of 0.9 and a learning rate of $0.025$. The architecture search takes about 1.4 GPU days on a single NVIDIA Tesla V100 (32G) GPU.

\vspace{\bfspace}
\noindent \textbf{DGS.} The DGS method is applied to the first stem layer (s1) during the retraining phase, we set $I_g = 100$, $\epsilon=0. 001$, $e_D=5$, $g=\mathrm{Dspike}$ with ${\{g_i\}}_N$ having equal temperature interval: $ \{g(b-\frac{(N-1)}{2}\Delta b),...,g(b+\frac{(N-1)}{2}\Delta b)\}$ where $\Delta b = 0.2$ and $N=5$. In extension experiments, we slightly increase the output channel of the first stem layer and use 3 nodes (n3) within a cell. In addition, we apply DGS to the first node of the 5th cell (c5) and the results can be found in the ablation study.

\vspace{\bfspace}
\noindent \textbf{TPS and HNS.} 
The search spaces for TPS and HNS are listed in Table \ref{tab:cla_search_space}.
To ensure stable training and prevent convergence to suboptimal solutions, we maintain fixed weighting factor $a$ and train the weights for a warm-up period of $5$ epochs before starting the search. 
We define each $100$ iterations as a search step, after each step $\tau$ is changed based on weighting factor $a$ and $a$ is reset for the next step. 

\vspace{\bfspace}
\noindent \textbf{Retrain.} After search, we retrain the model on target datasets with channel expansion for 100 epochs with mini-batch size $50$ for CIFAR and $160$ for ImageNet, with cosine learning rate 0.025. We use SGD optimizer with weight decay $3e^{-4}$ and momentum $0.9$. 

\subsubsection{Results}
The results are summarised in Table \ref{tab:cifar} and the values of other models are obtained from the literature. On the CIFAR, SpikeDHS-CLA outperforms most directly trained methods with similar model capacity. 
STTS \cite{yan2024sampling} reduces model parameters by introducing a light-weight depthwise separable convolution.
Among the pure spiking models, SpikeDHS-CLA with TPS achieves the highest accuracy. Note that both Dspike \cite{li2021differentiable} and TET \cite{deng2022temporal} use advanced training algorithms, while simple SpikeDHS-CLA models are trained with a fixed SG function.
Our model also outperforms recent SNN work with NAS in terms of accuracy, inference steps and model size\footnote{The size of SNASNet is not given.}.
The hybrid search further improves accuracy by introducing a small number of floating-point operations. In Section \ref{sec:tps_exp}, we provide a detailed analysis of the impact of different search spaces and search method settings of TPS on the results.

On ImageNet, we extand the SpikeDHS-CLA model to SpikeDHS-CLA-large by increasing from 8 layers to 10 layers. Our models outperform the ResNet-34 large model with a much smaller model capacity and compete with VGG-16 with less than half the size, as shown in Table \ref{tab:imagenet}. We speculate that the large number of jump connections, both within and across cells, potentially helps gradient propagation through deep layers, which improves model training.

\begin{table}[t!]
    \renewcommand{\arraystretch}{1.3}
    \caption{Search Space for SpikeDHS, TPS and DGS in stereo tasks.}
    \label{tab:stereo_search_space}
    \centering
        \begin{tabular}{cccc}\hline
        \noindent \textbf{SpikeDHS} &\noindent \textbf{DGS} &\noindent \textbf{TPS} &\noindent \textbf{HNS}  \\\hline

        \makecell[c]{conv3$\times$3 with $g(b=3)$\\conv3$\times$3 with $g(b=5)$\\skip connection} &
        \makecell[c]{$g_1$\\$g_2$\\...}  &
        \makecell[c]{$\text{LIF}(\tau)$\\ ALIF($\tau, \tau_a$)} &
        \makecell[c]{$\text{LIF}$\\ ALIF\\$\text{ReLU}$} 
        
        \\\bottomrule
    \end{tabular}
\end{table}

\subsection{Event-based Deep Stereo}

\begin{table*}[t!]
    \renewcommand{\arraystretch}{1.3}
    \caption{Comparison on MVSEC split 1. We denote the best and second best results in bold and underscore. EO denotes events-only method. EITNet requires gray scale images for training but not for inference. Symbols meaning: -, unavailability of the value; ${}^\mathrm{D}$: DGS method. ${}^\mathrm{T}$: TPS method. ${}^\mathrm{H}$: HNS method. NoP: Number of parameters. *, estimated value.}
    \label{tab:1-compare}
    \centering
    \begin{tabular}{ccccccccc}
        \toprule 
        Method 
        & EO 
        & NoP 
        & \makecell[c]{Inference\\speed} 
        & MDE {[}cm{]} ↓ 
        & \makecell[c]{Median depth\\error {[}cm{]} ↓} 
        & \makecell[c]{Mean disparity\\error {[}pix{]} ↓} 
        & 1PA {[}\%{]} ↑ \\
        \midrule
        EIS \cite{mostafavi2021event} & \ding{55} & - & - & \noindent \textbf{13.7} & - & - & 89.0 \\ 
        EITNet \cite{ahmed2021deep} & \checkmark\kern-1.1ex\raisebox{.7ex}{\rotatebox[origin=c]{125}{--}}  & $>$16M* & - & \underline{14.2} & \noindent \textbf{5.9} & 0.55 & \underline{92.1} \\ 
        \hdashline[1ex/2pt]
        DDES \cite{tulyakov2019learning} &\checkmark& 2.33M & 26 FPS& 16.7 & 6.8 & 0.59 & 89.4 \\
        StereoSpike \cite{ranccon2022stereospike} &\checkmark& -& - & 18.5 & - & - & - \\ 
        \noindent \textbf{SpikeDHS}  & \checkmark &0.87M &44 FPS& 15.7 & 6.3 & 0.55 & 91.0 \\ 
        $\noindent \textbf{SpikeDHS}^{\mathrm{D}}$ & \checkmark &0.87M &44 FPS &\text{15.4}& \underline{6.0} & \underline{0.54} & 91.3 \\ 
        $\noindent \textbf{SpikeDHS}^{\mathrm{T}}$ & \checkmark &0.87M &44 FPS &\text{15.2}& \noindent \textbf{5.9} & \underline{0.54} & 91.4 \\ 
        $\noindent \textbf{SpikeDHS}^{\mathrm{H}}$ & \checkmark &0.87M &44 FPS &\text{14.6}& \noindent \textbf{5.9} & \noindent \textbf{0.53} & \noindent \textbf{92.4} \\ 
        \bottomrule
        \end{tabular}
    \label{tab:mvsec}
\end{table*}

\begin{figure*}[t!]
    \centering
    \includegraphics[width=0.9\linewidth]{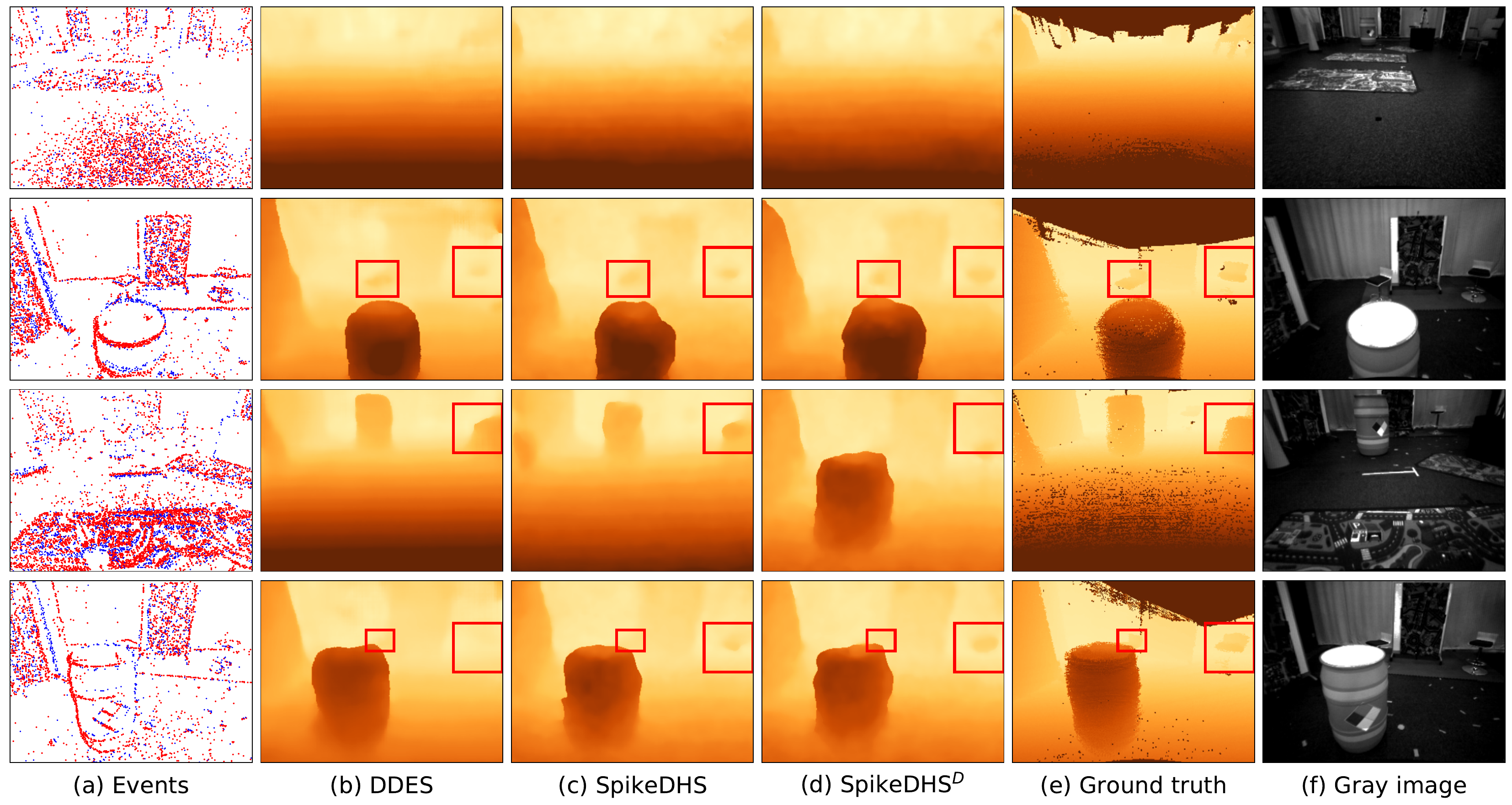}
    \caption{Qualitative comparison on MVSEC. Disparity maps of different methods are on same frames.}
    \label{fig:mvsec}
\end{figure*}

\subsubsection{Experimental Setup}

\noindent \textbf{Dataset.}  We further apply our method to event-based deep stereo matching on the widely used MVSEC dataset \cite{zhu2018multivehicle}.
The MVSEC dataset contains depth information recorded by LIDAR sensors and event streams collected by a pair of \emph{Davis346} cameras, with synchronized 20 Hz grayscale images at 346$\times$260 resolution. We split and preprocess the Indoor Flying dataset from the MVSEC using the same setup as \cite{tulyakov2019learning, ahmed2021deep, zhu2018realtime}. In split 1, 3110 samples from Indoor Flying 2 and 3 are used as the training set, while 861 and 200 samples from Indoor Flying 1 are used as the test and validation set, respectively. In split 3, 2600 samples from Indoor Flying 2 and 3 are used as the training set, while 1343 and 200 samples from Indoor Flying 1 are used as the test and validation set, respectively. For subsequent experiments, we utilize split 1, as the performance between splits 1 and 3 is nearly identical, with split 1 being the more commonly used configuration.

\vspace{\bfspace}
\noindent \textbf{SpikeDHS.} We use 3 nodes within a cell. 
The candidate operations are shown in the Table \ref{tab:stereo_search_space}.
For the layer search space, we adopt a four-level trellis with downsampling rates of \{3,2,2,2\}. 
The number of layers is set to 2 and 4 for the feature and matching net, respectively. Additionally, two stem layers are applied before both subnetworks to reduce the input spatial resolution and increase the number of channels.
We search the architecture for 12 epochs with batch size 1. The first 3 epochs are used to initiate the weight of the supernetwork. The remaining 9 epochs are applied with bi-level optimization. We use SGD optimizer with a momentum of $0.9$ and a learning rate of $0.002$. The architecture search takes about $0.4$ GPU days on a single NVIDIA Tesla V100 (32G) GPU.

\vspace{\bfspace}
\noindent \textbf{DGS.} The DGS method is applied to the first stem layer of the feature net with the same setup as in classification. 

\vspace{\bfspace}
\noindent \textbf{TPS and HNS.} 
For event-based stereo tasks, we search with both LIF and ALIF neurons. The overall search space is shown in Table \ref{tab:stereo_search_space}. We use the same TPS setting as for the classification tasks.

\vspace{\bfspace}
\noindent \textbf{Retrain.} After search, we retrain the model with channel expansion for 200 epochs with mini-batch size $2$. We use the Adam optimizer with an initial learning rate of $0.001$ and an impulse of $(0.9,0.999)$, where the learning rate is halved at $[50,100,150]$ epochs.

\subsubsection{Results}
\noindent \textbf{SpikeDHS.}
We compare our method with other event-based stereo matching approaches (Sec. \ref{sec:event_task}) on dense disparity estimation. The results are summarized in Table \ref{tab:1-compare}, values of other models are obtained from literature. Among SNN methods, SpikeDHS significantly outperforms StereoSpike, exhibiting the effectiveness of the searched architecture. Among events-only approaches, SpikeDHS even surpasses ANN-based specially designed DDES network in all criteria with only one-third of its number of parameters.

\vspace{\bfspace}
\noindent \textbf{DGS.} 
The performance of SpikeDHS-Stereo is further improved with DGS. As shown in Fig. \ref{fig:fig5}a-b, the temperature of the SG function is constantly optimized, which avoids vanishing gradient compared to training with fixed SG function, leading to more stable training of SNN.A qualitative comparison of estimated disparities is shown in Fig. \ref{fig:mvsec}. It can be seen that SpikeDHS with DGS predicts better disparities compares to other models, especially in local edges.

\vspace{\bfspace}
\noindent \textbf{TPS and HNS.} SpikeDHS-stereo with TPS exhibits a similar performance as DGS. Remarkably, using HNS the network even surpasses the ANN-based EITNet which requires grayscale images for training and has over 20 times more parameters, demonstrating the potential of integrating spiking and floating-point operations in event-based vision.

\vspace{\bfspace}
\noindent \textbf{Streaming inference.} In real-world scenario, events are generated consecutively by the sensor with flexible lengths. To test the real-time applicability of our model, we fed the entire test split continuously into the model, which evolves an equal length of steps and estimates sequential disparities. The inference speed of our model achieves 44 FPS (26 FPS for the DDES model) while achieving similar accuracy as in standard testing, where the model always receives a fixed length of events.
 
\subsection{Analysis on TPS and HNS}
\label{sec:tps_exp}
We perform a detailed study of the effectiveness of TPS and HNS. For TPS we evaluate its search process and influence on firing rate across layers. We also compare TPS and HNS with random selection, other handcrafted methods, and networks with fixed neuron types.

\subsubsection{Search process of TPS.} 
\begin{figure}[t!]
    \centering
    \subfigure[Search process on CIFAR100]{
        \includegraphics[width=0.45\linewidth]{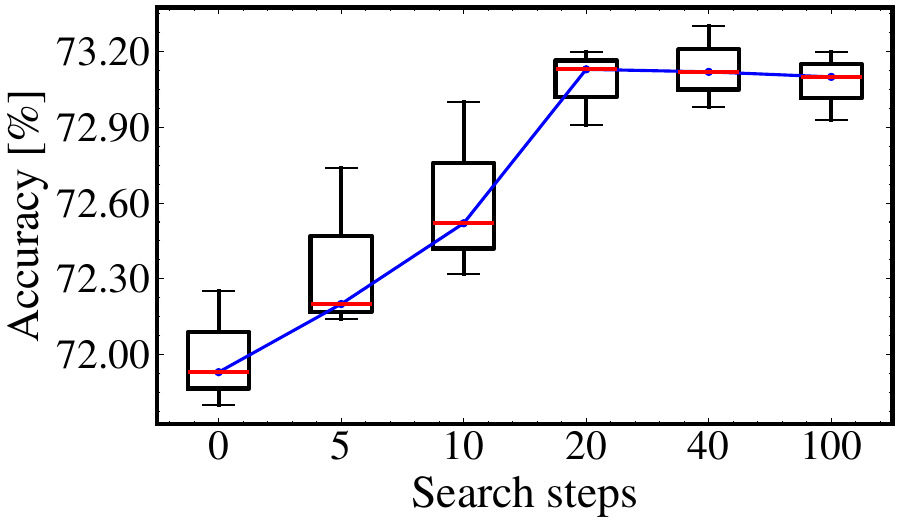}
        \label{fig:tps_cla}
    }
    \subfigure[Search process on MVSEC]{
        \includegraphics[width=0.45\linewidth]{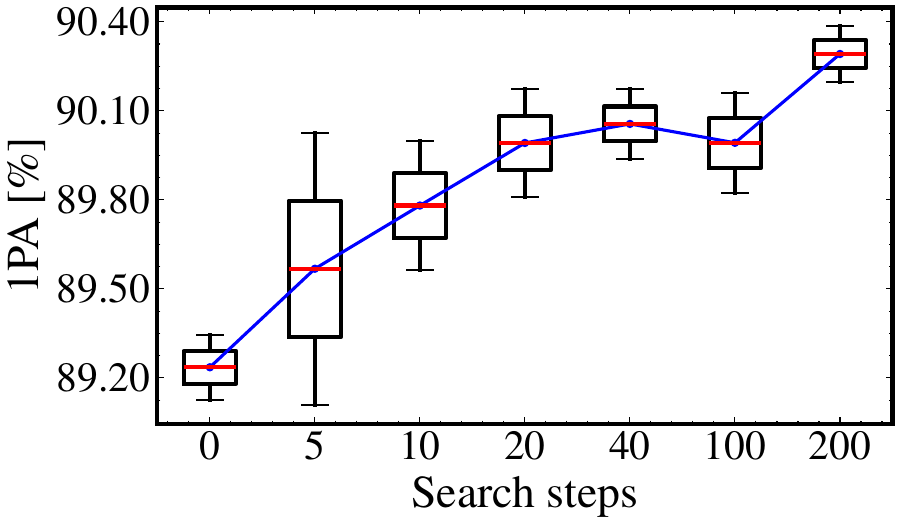}
        \label{fig:tps_stereo}
    }
    \caption{Search process of TPS on (a) CIFAR100 (b) MVSEC dataset. Each snapshot is retrained from scratch for fewer epochs and evaluated on the validation set. The results are averaged over multiple experiments.}
    \label{fig:tps_searchprocess}
\end{figure}

Fig. \ref{fig:tps_searchprocess} evaluates the search process of TPS on CIFAR100 and MVSEC tasks. Specifically, throughout the search process, we keep track of the most recent optimal hyperparameters over time. Each checkpoint is retrained for 100 epochs (CIFAR100) or 40 epochs (MVSEC) and then evaluated using the test set. We repeat the experiments three times with different random seeds. Notably, the performance gradually improves as the search progresses. This clear improvement demonstrates the effectiveness of TPS in optimizing membrane time constants.

\begin{table}[t!]
    \renewcommand{\arraystretch}{1.3}
    \caption{Comparison of TPS and HNS on image classification on CIFAR100. Note that the model architecture is obtained from SpikeDHS. Bold values are using TPS or HNS select method. With $(\tau)$ means that $\tau$ can be changed.}
    \label{tab:tps_cla_performance}
    \centering
\begin{tabular}{cccc}
    \hline
    Neurons & Select Method & Energy & Auucracy \\
    \hline
    LIF & Handcrafted & 5.5mJ & 76.15 \\
    
    LIF($\tau$) & Direct gradient & 5.6mJ & 76.53 \\ 
    
    PLIF & Direct gradient & 5.7mJ & 77.8 \\ 
    
    GLIF & Direct gradient & 5.5mJ & 76.94 \\ 
    
    ALIF & Handcrafted & 5.3mJ & 75.21 \\
    
    ReLU & Handcrafted & 27.6mJ & 79.25 \\ 
    \hline
    LIF($\tau$) & Random & 5.8mJ & 76.62 \\ 
    
    LIF($\tau$) & TPS & 5.6mJ & {\bf 77.9} \\ 
    
    LIF,ReLU & HNS w/ $\mathcal{L}_{\text{eg}}$ & 13.2mJ & {\bf 78.3} \\ 
    
    LIF,ReLU & HNS & 18.7mJ & {\bf 78.96} \\ 
    \hline
    \end{tabular}     
\end{table}
   
\begin{table}[!t]
    \renewcommand{\arraystretch}{1.1}
    \caption{Comparison of TPS and HNS on event-based stereo on MVSEC. Note that the model architecture is obtained from SpikeDHS. Bold values are using TPS or HNS select method. With $(\tau)$ means that $\tau$ can be changed.}
    \label{tab:tps_mvsec}
    \centering
    \footnotesize
\begin{tabular}{cccc}
    \hline
    Neurons & Select Method & Energy & 1PA {[}\%{]} ↑  \\
    \hline
    LIF & Handcrafted & 6.7mJ & 90.14 \\
    
    LIF($\tau$) & Direct gradient & 6.6mJ & 90.60 \\
    
    PLIF & Direct gradient & 6.7mJ & 90.64 \\
    
    GLIF & Direct gradient & 6.4mJ & 90.76 \\
    
    ALIF & Handcrafted & 6.6mJ & 91.28 \\
    
    ALIF($\tau$) & Direct gradient & 6.6mJ & 90.54 \\
    
    ALIF($\tau,\tau_a$) & Direct gradient & 6.6mJ & 90.53 \\
    
    IF & Handcrafted & 6.4mJ & 88.30 \\
    
    ReLU & Handcrafted & 34.1mJ & 92.22 \\
    \hline
    LIF($\tau$) & Random & 6.8mJ & 90.12 \\
    
    LIF($\tau$) & TPS & 6.7mJ & {\bf 91.19} \\
    
    ALIF($\tau, \tau_a$) & TPS & 6.7mJ & {\bf 91.34} \\
    
    LIF,ALIF,ReLU & HNS w/ $\mathcal{L}_{\text{eg}}$ & 17.6mJ & {\bf 91.83} \\
    
    LIF,ALIF,ReLU & HNS & 23.1mJ & {\bf 92.42} \\
    \hline
    \end{tabular}

    \end{table}

    \begin{figure*}[t!]
        \centering
        \subfigure[Network firing rate over search steps]{
            \includegraphics[width=0.315\linewidth]{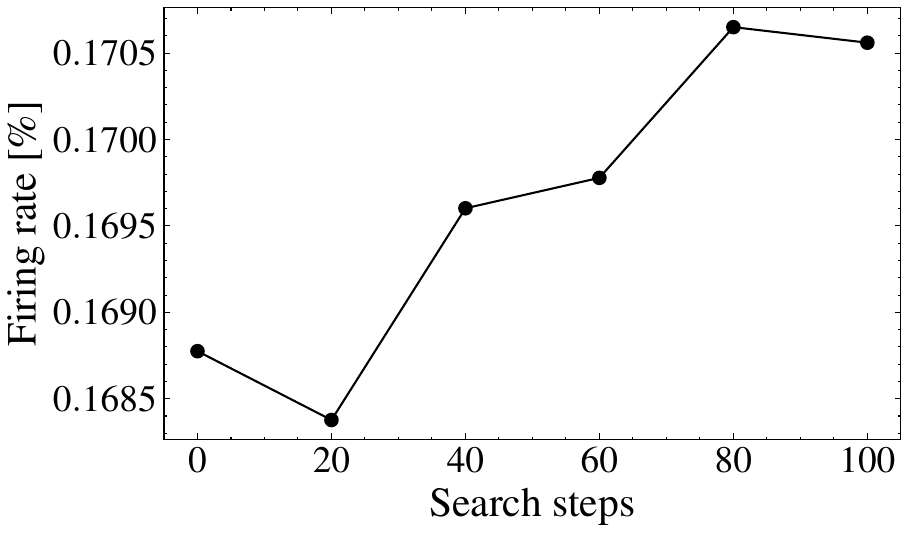}
            \label{fig:tps_fr_step}
        }
        \subfigure[Layer-wise firing rate]{
            \includegraphics[width=0.315\linewidth]{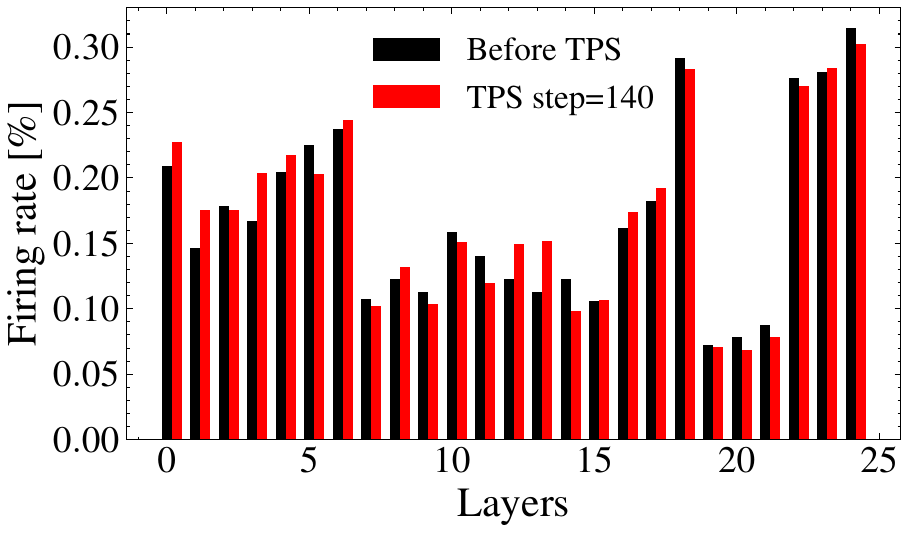}
            \label{fig:tps_fr_layer}
        }
        \subfigure[$\tau$ changes over search steps]{
            \includegraphics[width=0.315\linewidth]{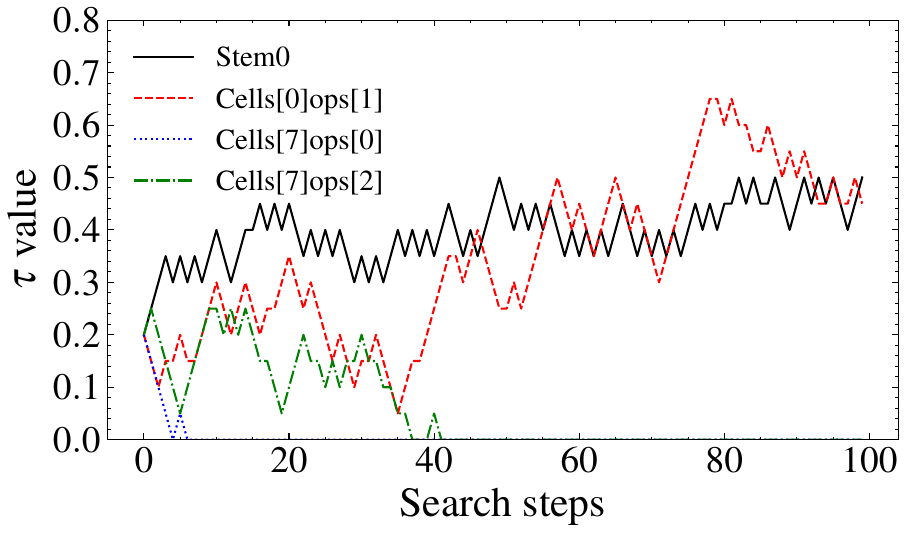}
            \label{fig:tps_tau_step}
        }
        \caption{Detailed analysis of the TPS process. (a) The average network firing rate increases gradually over the search steps, suggesting stronger information representation. (b) The layer-wise firing rate, before and after TPS, shows a non-uniform change across layers: earlier layers tend to have higher firing rates, while later layers see a decrease. This indicates TPS adjusts the activation levels differently across the network. (c) Changes in $\tau$ show an increase in the stem and a decrease in some later layers, reflecting varying needs for temporal information.}
        \label{fig:tps_prove}
    \end{figure*}

\subsubsection{Comparison of TPS with other methods.} 
We conduct comparative analysis of TPS with random selection, handcrafted designs, and the direct gradient method on the CIFAR100 and MVSEC datasets. 
In the random selection approach, $\tau$ is randomly assigned a value between 0 and 1 multiple times, and the average accuracy over all training iterations is computed. For the handcrafted design, $\tau$ or $\tau_a$ is fixed at the default value of 0.2.
The results are presented in Table \ref{tab:tps_cla_performance} and Table \ref{tab:tps_mvsec}. 
The results demonstrate the consistent superiority of the TPS method over other methods. In particular, LIF($\tau$), PLIF and GLIF use direct gradient training methods, the difference between direct gradient traning and TPS is mentioned in section \ref{sec:tps}. 

\subsubsection{Comparison of HNS with networks of fixed neuron.}
We further compare HNS with fixed neuron, it can be clearly seem in Table \ref{tab:tps_cla_performance} and Table \ref{tab:tps_mvsec} that inclusion of certain floating-point operation can indeed improve the performance for both CIFAR100 and MVSEC datasets. 
Notably, on MVSEC dataset, HNS outperforms the ReLU-only network in terms of both accuracy and energy consumption. The results highlight the potential of integrating spiking and floating-point neurons for event-based vision tasks rich in both temporal and spatial information.
Futhermore, we integrate the energy loss function $L_\text{eg}$ to limits to energy consumption.
The proposed energy loss achieves a balanced trade-off between energy efficiency and accuracy.
 
\subsubsection{Firing rate analysis.} 
We further propose the question: what factors lead to the improvement in performance? 
One potential reason is the optimization of membrane time constants, which affects the distribution of membrane potential and subsequently the firing distribution. By comparing the firing rate on CIFAR100 before and after application of TPS, indeed we observe a noticeable difference. We plot the average firing rate of the whole network at each snapshot resulting from the search process, as shown in Fig. \ref{fig:tps_fr_step}. The firing rate slightly increases as the search steps progress, indicating a strengthening of information representation. 
We then examine the firing rate distribution across layers. As shown in Fig. \ref{fig:tps_fr_layer}. notably, the firing rate does not increase uniformly across all layers; instead, certain layers experience a decrease. Specifically, the firing rate mainly increases at front layers, while decreases for most later layers. The observation demonstrates a dynamic modulation of TPS on activation for different parts of the network.

To further investigate this phenomenon, we plot the evolution of membrane time constants $\tau$ for several front and later layers as the search progresses. As shown in Fig. \ref{fig:tps_tau_step}, $\tau$ increases in the stem and the front cell layers, while it decreases in the later cell layers.
This observation suggests a heterogeneous demand of temporal information within the network. By modulating temporal parameters locally, TPS enables the SNN to exploit diverse temporal dynamics, leading to an improvement in performance.

\subsection{Energy Cost Estimation}
\begin{figure*}[t!]
    \centering
    \includegraphics[width=.95\linewidth]{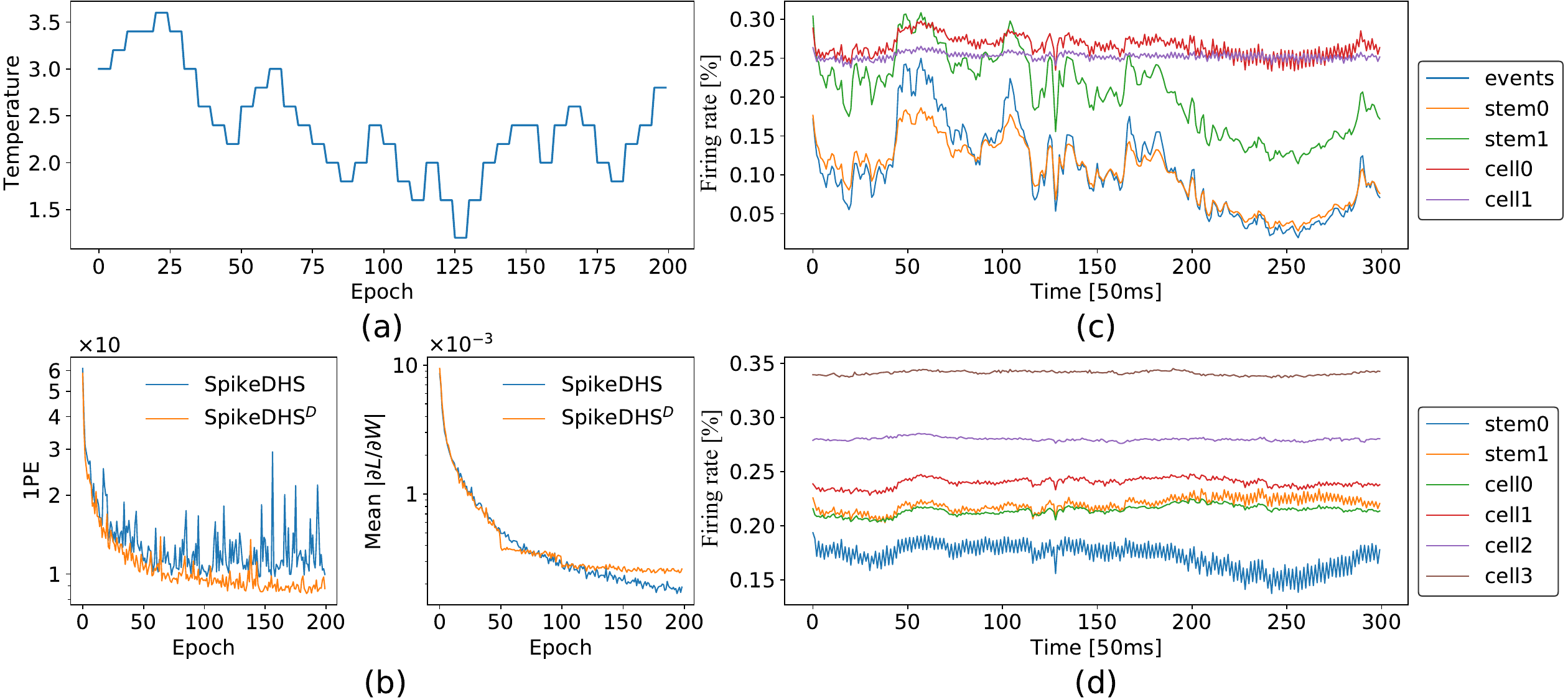}
    \caption{DGS training process and firing rate of SpikeDHS-Stereo. (a): Evolution of SG temperatures during DGS training. (b): 1PE (left) and first stem weight gradient (right) of DGS training and normal training. (c) and (d): Firing rate of the feature net and the matching net.} 
    \label{fig:fig5}
\end{figure*}

The dense computation of deep ANNs came at a significant energy cost. In contrast, SNNs perform sparse computation and multiplication-free inference. As shown in Fig. \ref{fig:tps_fr_layer}, our models show an overall sparse activity with different degrees across layers. For event-based stereo, we plot the network activity together with the density of events for a short duration (Fig. \ref{fig:fig5}c-d). The activity of the first stem layer in the feature network is highly correlated with the density of event streams. This relationship weakens with increasing layer depth. As the activity propagates, the firing rate of matching layers decreases to create dense disparities. The ANN firing rate is collected by a publicly available PyTorch package \cite{ptflops}. After \cite{li2021differentiable}, the addition number of SNN is calculated by $\mathop{fr}*T*A$, where $\mathop{fr}$ is the mean firing rate, $T$ is the time step, and $A$ is the addition number. The number of operations and the energy cost are in the Table \ref{tab:energy}. We can see that SpikeDHS-Stereo has a much lower number of operations than DDES. Note that due to the streaming inference, SNN realizes a natural use of its temporal dynamics with $T=1$. We measure the energy consumption after \cite{rathi2021diet}. In $45 \mathrm{nm}$ CMOS technology, the addition operation in SNN costs $0.9\mathrm{pJ}$, while the MAC operation in ANN consumes $4.6\mathrm{pJ}$. SpikeDHS-Stereo costs only $6.7\mathrm{mJ}$ for a single forward, with $26\times$ less energy than DDES.

\begin{table}[t!]
\centering
\fontsize{7.7pt}{\baselineskip}\selectfont 
\setlength\tabcolsep{2.4pt}
\caption{The operation number and energy cost.}
\begin{tabular}{lrrrrr}
\toprule
\multicolumn{1}{l}{Method} &
\multicolumn{1}{l}{\#Add.} &
\multicolumn{1}{l}{\#Mult.} &
\multicolumn{1}{l}{Energy}\\ 
\midrule
DARTS CIFAR10   & 29654M & 29654M & 136.4 $\mathrm{mJ}$ \\
SpikeDHS-CLA (n4) & \noindent \textbf{5973M} & \noindent \textbf{19M} & \noindent \textbf{5.5 $\mathrm{mJ}$} \\
\midrule
DDES (ANN)    & 37917M &37917M & 174.4 $\mathrm{mJ}$\\
SpikeDHS-Stereo & \noindent \textbf{7414M} &	\noindent \textbf{0M} &	\noindent \textbf{6.7 $\mathrm{mJ}$}  \\
\bottomrule
\end{tabular}
\label{tab:energy}
\end{table}

\section{Ablation Study}

\subsection{Operation Mixing Position}
As mentioned in Sec. \ref{sec:cell_search}, during the search we can mix the operation either at the spike activation or at the membrane potential. The comparison results are shown in Table \ref{tab:alb_mvsec}. 
The performance of mixing at membrane is better than mixing at spike activation. One potential reason is mixing at spike activation should pass additional step function which introduce additional SG function prediction error.

\subsection{Temporal Dynamics in Event-Based Task}
The inherent temporal dynamics of SNN is thought to be helpful in learning the temporal correlation of the data. To investigate this property, we fix the membrane time constant $\tau$ to 0 to create a binary neuron (BI) and perform architecture search and retraining. As shown in Table \ref{tab:mvsec}, the SpikeDHS-BI model performs worse than SpikeDHS with dynamic neurons, demonstrating the benefit of temporal dynamics in this task.

\begin{table}[t!]
    \renewcommand{\arraystretch}{1.3}
    \caption{Spiking activation position and temporal dynamics in MVSEC split1. MM denotes mix at the input on the membrane potential while MS denotes mix at the spike activation. BI denotes using binary neuron.}
    \label{tab:alb_mvsec}
    \centering
    \begin{tabular}{cccccccc}
    \toprule 
    \multirow{1}{*}{Method} 
    & \multicolumn{1}{c}{MDE {[}cm{]} ↓} 
    & \multicolumn{1}{c}{1PA {[}\%{]} ↑} \\
    \midrule
    \text{SpikeDHS (MM)}  & 15.7 &91.0  \\
    \text{SpikeDHS (MS)} &  16.5 & 90.1 \\
    SpikeDHS (BI)  &17.5  & 88.7  \\ 
    \bottomrule
    \end{tabular}
    \label{tab:mvsec}
\end{table}

\subsection{SpikeDHS Node Number and DGS location}
To analyze the influence of node number and DGS location, we conduct comparison experiments with 3 and 4 nodes in a cell, as well as DGS in stem layer (s1) or the first node of the 5th cell (c5), the results are shown in Table \ref{tab:cla_node_dgs}. 3 nodes perform better than 4 nodes and DGS in different position seems change slightly.

\begin{table}[t!]
    \renewcommand{\arraystretch}{1.3}
    \centering
    \caption{SpikeDHS node number and DGS position influence on CIFAR. ${}^\mathrm{D}$: DGS method. NoP: Number of parameters.}
    \label{tab:cla_node_dgs}
    \begin{tabular}{cccc}
    \toprule Architecture  & NoP  & CIFAR10 & CIFAR100 \\
    \midrule
    n4  & 12M  & 94.34 & 75.70 \\
    n3 & 14M  & 95.35  &76.15 \\
    n4s1 (DGS) & 12M  & 94.68  & 76.03 \\
    n3s1 (DGS) & 14M  & 95.36  & 76.25 \\
    n3c5 (DGS) & 14M  & 95.50  & 76.23 \\
    \bottomrule
\end{tabular}
\end{table}

\subsection{Different SGs for DGS}

\begin{figure}[!th]
    \centering
    \includegraphics[width=0.85\linewidth]{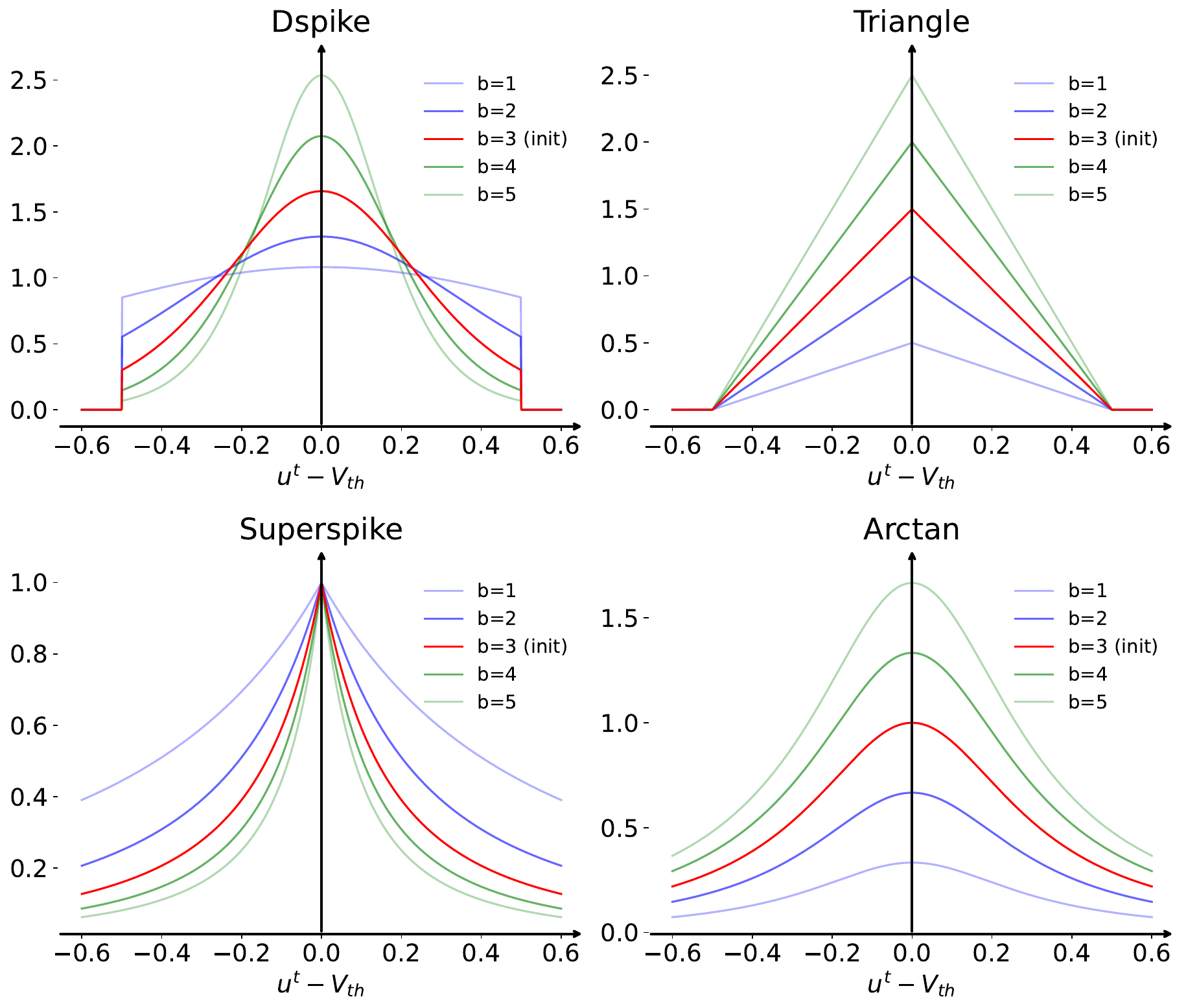}
    \caption{The shape of different SG functions with different temperature factor $b$. We set the initial value of $b$ to $3$.}
    \label{fig:diffsg}
  \end{figure}
  
To test the robustness of DGS for different SG functions, we evaluate four different SG functions including Dspike \cite{li2021differentiable} (Eq. \ref{eq:dspike}), Triangle \cite{bellec2018long} (Eq. \ref{eq:triangle}), Superspike \cite{zenke2018superspike} (Eq. \ref{eq:superspike}) and Arctan \cite{fang2021incorporating} (Eq. \ref{eq:arctan}) functions with fixed hyperparameters during training and varying hyperparameters with DGS. These functions are described as following: 
\begin{equation}
\delta^{\prime}_D(x)= \frac{b}{2} \cdot \frac{1-\mathrm{tanh}^2(b(x-\frac{1}{2}))}{\mathrm{tanh}(\frac{b}{2})} \text{ if } 0\leq x \leq 1
\label{eq:dspike}
\end{equation}
\begin{equation}
\delta^{\prime}_T(x)= b \cdot \mathrm{max}\{0,1-|x-\frac{1}{2}|\}
\label{eq:triangle}
\end{equation}
\begin{equation}
\delta^{\prime}_S(x)= \frac{1}{(b \cdot |x|+1)^2}
\label{eq:superspike}
\end{equation}
\begin{equation}
\delta^{\prime}_A(x)= \frac{b/3}{1+(\pi x)^2}
\label{eq:arctan}
\end{equation}
Each SG function has a temperature factor $b$ to control its shape through DGS. A visualization is shown in Fig. \ref{fig:diffsg}. The results demonstrate the robustness of DGS for different SG functions as it consistently improves network performance, as shown in Table \ref{tab:mvsec-diff_SG}.
       
\begin{table}[t!]
    \caption{Different SG functions for event-based stereo task on split 1 w/ and w/o DGS.}
    \label{tab:mvsec-diff_SG}
    \renewcommand{\arraystretch}{1.3}
    \centering
        \begin{tabular}{cc}\hline
        \noindent \textbf{SG function} &\noindent \textbf{1PA w/ DGS (w/o DGS) {[}\%{]} ↑}\\\hline
        Triangle \cite{bellec2018long} & 91.3 (90.9) \\
        Arctan \cite{fang2021incorporating} & 90.1 (89.3) \\
        Superspike \cite{zenke2018superspike} & 89.7 (89.6) \\\hline
        \end{tabular}
    \label{tab:cla}
    \end{table}

\section{Conclusion}
\label{sec-Conclusion}
In this work, we propose a spatial-temporal search methodology for optimizing SNNs that considers both architectural and temporal dimensions. The SpikeDHS framework finds optimal network architectures for specific tasks with limited computational cost by exploiting spike-based operations. The DGS method further improves the efficiency of gradient approximation during SNN training. In the temporal domain, inspired by the diversity of neural dynamics, we develop TPS to discover the optimal temporal dynamics of SNN, which further enables the exploration of efficient hybrid networks.
In image classification and event-based deep stereo tasks, our approach demonstrates competitive performance with state-of-the-art SNNs and ANNs. In particular, on event-based deep stereo tasks, our SNN outperforms even custom-designed ANNs in terms of both accuracy and computational cost, the optimized hybrid network extends the performance even further, demonstrating the potential of integrating dynamic spiking and static floating-point operations in event-based vision.
Our method can be applied to other tasks where networks require more architectural variations, such as object detection and semantic segmentation. Currently, we use only a limited search space; other types of biologically plausible connections, such as recurrent and feedback connections, as well as excitatory/inhibitory synapses and neurons with more complex dynamics could be considered in the future.


\section*{Acknowledgements}
The study was funded by the National Natural Science Foundation of China under contracts No. 62332002, No. 62027804, No. 61825101, No.62206141, No.62236009, the major key project of the Peng Cheng Laboratory (PCL2021A13), and Shenzhen Basic Research Program under Grant JCYJ20220813151736001. Computing support was provided by Pengcheng Cloudbrain.
\ifCLASSOPTIONcaptionsoff
  \newpage
\fi



\bibliographystyle{IEEEtran}
%
\bibliography{bib.bib}

\begin{thebibliography}{100}
\providecommand{\url}[1]{#1}
\csname url@samestyle\endcsname
\providecommand{\newblock}{\relax}
\providecommand{\bibinfo}[2]{#2}
\providecommand{\BIBentrySTDinterwordspacing}{\spaceskip=0pt\relax}
\providecommand{\BIBentryALTinterwordstretchfactor}{4}
\providecommand{\BIBentryALTinterwordspacing}{\spaceskip=\fontdimen2\font plus
\BIBentryALTinterwordstretchfactor\fontdimen3\font minus \fontdimen4\font\relax}
\providecommand{\BIBforeignlanguage}[2]{{%
\expandafter\ifx\csname l@#1\endcsname\relax
\typeout{** WARNING: IEEEtran.bst: No hyphenation pattern has been}%
\typeout{** loaded for the language `#1'. Using the pattern for}%
\typeout{** the default language instead.}%
\else
\language=\csname l@#1\endcsname
\fi
#2}}
\providecommand{\BIBdecl}{\relax}
\BIBdecl

\bibitem{maass1997networks}
W.~Maass, ``Networks of spiking neurons: the third generation of neural network models,'' \emph{Neural Networks}, vol.~10, no.~9, pp. 1659--1671, 1997.

\bibitem{bohte2000spikeprop}
S.~M. Bohte, J.~N. Kok, and J.~A. La~Poutr{\'e}, ``Spikeprop: backpropagation for networks of spiking neurons.'' in \emph{European Symposium on Artificial Neural Networks (ESANN)}, vol.~48.\hskip 1em plus 0.5em minus 0.4em\relax Bruges, 2000, pp. 419--424.

\bibitem{wu2018spatio}
Y.~Wu, L.~Deng, G.~Li, J.~Zhu, and L.~Shi, ``Spatio-temporal backpropagation for training high-performance spiking neural networks,'' \emph{Frontiers in Neuroscience}, vol.~12, p. 331, 2018.

\bibitem{neftci2019surrogate}
E.~O. Neftci, H.~Mostafa, and F.~Zenke, ``Surrogate gradient learning in spiking neural networks: Bringing the power of gradient-based optimization to spiking neural networks,'' \emph{IEEE Signal Processing Magazine}, vol.~36, no.~6, pp. 51--63, 2019.

\bibitem{shrestha2018slayer}
S.~B. Shrestha and G.~Orchard, ``Slayer: Spike layer error reassignment in time,'' \emph{Proceedings of Advances in Neural Information Processing Systems (NeurIPS)}, vol.~31, 2018.

\bibitem{wu2019direct}
Y.~Wu, L.~Deng, G.~Li, J.~Zhu, Y.~Xie, and L.~Shi, ``Direct training for spiking neural networks: Faster, larger, better,'' in \emph{Proceedings of Association for the Advancement of Artificial Intelligence (AAAI)}, vol.~33, no.~01, 2019, pp. 1311--1318.

\bibitem{zhang2020temporal}
W.~Zhang and P.~Li, ``Temporal spike sequence learning via backpropagation for deep spiking neural networks,'' \emph{Proceedings of Advances in Neural Information Processing Systems (NeurIPS)}, vol.~33, pp. 12\,022--12\,033, 2020.

\bibitem{rathi2021diet}
N.~Rathi and K.~Roy, ``Diet-snn: A low-latency spiking neural network with direct input encoding and leakage and threshold optimization,'' \emph{IEEE Transactions on Neural Networks and Learning Systems}, 2021.

\bibitem{zheng2021going}
H.~Zheng, Y.~Wu, L.~Deng, Y.~Hu, and G.~Li, ``Going deeper with directly-trained larger spiking neural networks,'' in \emph{Proceedings of Association for the Advancement of Artificial Intelligence (AAAI)}, vol.~35, no.~12, 2021, pp. 11\,062--11\,070.

\bibitem{he2016deep}
K.~He, X.~Zhang, S.~Ren, and J.~Sun, ``Deep residual learning for image recognition,'' in \emph{Proceedings of the IEEE/CVF Conference on Computer Vision and Pattern Recognition (CVPR)}, 2016, pp. 770--778.

\bibitem{simonyan2014very}
K.~Simonyan and A.~Zisserman, ``Very deep convolutional networks for large-scale image recognition,'' \emph{arXiv preprint arXiv:1409.1556}, 2014.

\bibitem{li2021differentiable}
Y.~Li, Y.~Guo, S.~Zhang, S.~Deng, Y.~Hai, and S.~Gu, ``Differentiable spike: Rethinking gradient-descent for training spiking neural networks,'' \emph{Proceedings of Advances in Neural Information Processing Systems (NeurIPS)}, vol.~34, 2021.

\bibitem{fang2021deep}
W.~Fang, Z.~Yu, Y.~Chen, T.~Huang, T.~Masquelier, and Y.~Tian, ``Deep residual learning in spiking neural networks,'' \emph{Proceedings of Advances in Neural Information Processing Systems (NeurIPS)}, vol.~34, 2021.

\bibitem{deng2022temporal}
S.~Deng, Y.~Li, S.~Zhang, and S.~Gu, ``Temporal efficient training of spiking neural network via gradient re-weighting,'' \emph{arXiv preprint arXiv:2202.11946}, 2022.

\bibitem{zhu2022event}
L.~Zhu, X.~Wang, Y.~Chang, J.~Li, T.~Huang, and Y.~Tian, ``Event-based video reconstruction via potential-assisted spiking neural network,'' in \emph{Proceedings of the IEEE/CVF Conference on Computer Vision and Pattern Recognition (CVPR)}, 2022, pp. 3594--3604.

\bibitem{hagenaars2021self}
J.~Hagenaars, F.~Paredes-Vall{\'e}s, and G.~De~Croon, ``Self-supervised learning of event-based optical flow with spiking neural networks,'' \emph{Proceedings of Advances in Neural Information Processing Systems (NeurIPS)}, vol.~34, 2021.

\bibitem{kim2022beyond}
Y.~Kim, J.~Chough, and P.~Panda, ``Beyond classification: Directly training spiking neural networks for semantic segmentation,'' \emph{Neuromorphic Computing and Engineering}, vol.~2, no.~4, p. 044015, 2022.

\bibitem{mountcastle1997columnar}
V.~B. Mountcastle, ``The columnar organization of the neocortex.'' \emph{Brain: a Journal of Neurology}, vol. 120, no.~4, pp. 701--722, 1997.

\bibitem{liang2017interactions}
H.~Liang, X.~Gong, M.~Chen, Y.~Yan, W.~Li, and C.~D. Gilbert, ``Interactions between feedback and lateral connections in the primary visual cortex,'' \emph{Proceedings of the National Academy of Sciences}, vol. 114, no.~32, pp. 8637--8642, 2017.

\bibitem{elsken2019neural}
T.~Elsken, J.~H. Metzen, and F.~Hutter, ``Neural architecture search: A survey,'' \emph{The Journal of Machine Learning Research}, vol.~20, no.~1, pp. 1997--2017, 2019.

\bibitem{liu2021survey}
Y.~Liu, Y.~Sun, B.~Xue, M.~Zhang, G.~G. Yen, and K.~C. Tan, ``A survey on evolutionary neural architecture search,'' \emph{IEEE Transactions on Neural Networks and Learning Systems}, 2021.

\bibitem{zoph2018learning}
B.~Zoph, V.~Vasudevan, J.~Shlens, and Q.~V. Le, ``Learning transferable architectures for scalable image recognition,'' in \emph{Proceedings of the IEEE/CVF Conference on Computer Vision and Pattern Recognition (CVPR)}, 2018, pp. 8697--8710.

\bibitem{zoph2016neural}
B.~Zoph and Q.~V. Le, ``Neural architecture search with reinforcement learning,'' \emph{arXiv preprint arXiv:1611.01578}, 2016.

\bibitem{liu2019auto}
C.~Liu, L.-C. Chen, F.~Schroff, H.~Adam, W.~Hua, A.~L. Yuille, and L.~Fei-Fei, ``Auto-deeplab: Hierarchical neural architecture search for semantic image segmentation,'' in \emph{Proceedings of the IEEE/CVF Conference on Computer Vision and Pattern Recognition (CVPR)}, 2019, pp. 82--92.

\bibitem{liudarts}
H.~Liu, K.~Simonyan, and Y.~Yang, ``Darts: Differentiable architecture search,'' in \emph{Proceedings of International Conference on Learning Representations (ICLR)}, 2018.

\bibitem{hasselmo2006mechanisms}
M.~E. Hasselmo and C.~E. Stern, ``Mechanisms underlying working memory for novel information,'' \emph{Trends in Cognitive Sciences}, vol.~10, no.~11, pp. 487--493, 2006.

\bibitem{shankar2012scale}
K.~H. Shankar and M.~W. Howard, ``A scale-invariant internal representation of time,'' \emph{Neural Computation}, vol.~24, no.~1, pp. 134--193, 2012.

\bibitem{kandel2000principles}
E.~R. Kandel, J.~H. Schwartz, T.~M. Jessell, S.~Siegelbaum, A.~J. Hudspeth, S.~Mack \emph{et~al.}, \emph{Principles of neural science}.\hskip 1em plus 0.5em minus 0.4em\relax McGraw-hill New York, 2000, vol.~4.

\bibitem{gerstner2014neuronal}
W.~Gerstner, W.~M. Kistler, R.~Naud, and L.~Paninski, \emph{Neuronal dynamics: From single neurons to networks and models of cognition}.\hskip 1em plus 0.5em minus 0.4em\relax Cambridge University Press, 2014.

\bibitem{zhu2018multivehicle}
A.~Z. Zhu, D.~Thakur, T.~{\"O}zaslan, B.~Pfrommer, V.~Kumar, and K.~Daniilidis, ``The multivehicle stereo event camera dataset: An event camera dataset for 3d perception,'' \emph{IEEE Robotics and Automation Letters}, vol.~3, no.~3, pp. 2032--2039, 2018.

\bibitem{petrovici2016stochastic}
M.~A. Petrovici, J.~Bill, I.~Bytschok, J.~Schemmel, and K.~Meier, ``Stochastic inference with spiking neurons in the high-conductance state,'' \emph{Physical Review E}, vol.~94, no.~4, p. 042312, 2016.

\bibitem{neftci2014event}
E.~Neftci, S.~Das, B.~Pedroni, K.~Kreutz-Delgado, and G.~Cauwenberghs, ``Event-driven contrastive divergence for spiking neuromorphic systems,'' \emph{Frontiers in Neuroscience}, vol.~7, p. 272, 2014.

\bibitem{leng2018spiking}
L.~Leng, R.~Martel, O.~Breitwieser, I.~Bytschok, W.~Senn, J.~Schemmel, K.~Meier, and M.~A. Petrovici, ``Spiking neurons with short-term synaptic plasticity form superior generative networks,'' \emph{Scientific Reports}, vol.~8, no.~1, pp. 1--11, 2018.

\bibitem{mozafari2018first}
M.~Mozafari, S.~R. Kheradpisheh, T.~Masquelier, A.~Nowzari-Dalini, and M.~Ganjtabesh, ``First-spike-based visual categorization using reward-modulated stdp,'' \emph{IEEE Transactions on Neural Networks and Learning Systems}, vol.~29, no.~12, pp. 6178--6190, 2018.

\bibitem{leng2020solving}
L.~Leng, ``Solving machine learning problems with biological principles,'' Ph.D. dissertation, 2020.

\bibitem{korcsak2022cortical}
A.~Korcsak-Gorzo, M.~G. M{\"u}ller, A.~Baumbach, L.~Leng, O.~J. Breitwieser, S.~J. van Albada, W.~Senn, K.~Meier, R.~Legenstein, and M.~A. Petrovici, ``Cortical oscillations support sampling-based computations in spiking neural networks,'' \emph{PLoS Computational Biology}, vol.~18, no.~3, p. e1009753, 2022.

\bibitem{bellec2020solution}
G.~Bellec, F.~Scherr, A.~Subramoney, E.~Hajek, D.~Salaj, R.~Legenstein, and W.~Maass, ``A solution to the learning dilemma for recurrent networks of spiking neurons,'' \emph{Nature Communications}, vol.~11, no.~1, pp. 1--15, 2020.

\bibitem{rueckauer2017conversion}
B.~Rueckauer, I.-A. Lungu, Y.~Hu, M.~Pfeiffer, and S.-C. Liu, ``Conversion of continuous-valued deep networks to efficient event-driven networks for image classification,'' \emph{Frontiers in Neuroscience}, vol.~11, p. 682, 2017.

\bibitem{bu2021optimal}
T.~Bu, W.~Fang, J.~Ding, P.~Dai, Z.~Yu, and T.~Huang, ``Optimal ann-snn conversion for high-accuracy and ultra-low-latency spiking neural networks,'' in \emph{Proceedings of International Conference on Learning Representations (ICLR)}, 2021.

\bibitem{li2021free}
Y.~Li, S.~Deng, X.~Dong, R.~Gong, and S.~Gu, ``A free lunch from ann: Towards efficient, accurate spiking neural networks calibration,'' in \emph{Proceedings of International Conference on Machine Learning (ICML)}.\hskip 1em plus 0.5em minus 0.4em\relax PMLR, 2021, pp. 6316--6325.

\bibitem{wu2021progressive}
J.~Wu, C.~Xu, X.~Han, D.~Zhou, M.~Zhang, H.~Li, and K.~C. Tan, ``Progressive tandem learning for pattern recognition with deep spiking neural networks,'' \emph{IEEE Transactions on Pattern Analysis and Machine Intelligence}, vol.~44, no.~11, pp. 7824--7840, 2021.

\bibitem{bengio2013estimating}
Y.~Bengio, N.~L{\'e}onard, and A.~Courville, ``Estimating or propagating gradients through stochastic neurons for conditional computation,'' \emph{arXiv preprint arXiv:1308.3432}, 2013.

\bibitem{yin2019understanding}
P.~Yin, J.~Lyu, S.~Zhang, S.~Osher, Y.~Qi, and J.~Xin, ``Understanding straight-through estimator in training activation quantized neural nets,'' in \emph{Proceedings of International Conference on Learning Representations (ICLR)}, 2019.

\bibitem{zenke2021remarkable}
F.~Zenke and T.~P. Vogels, ``The remarkable robustness of surrogate gradient learning for instilling complex function in spiking neural networks,'' \emph{Neural Computation}, vol.~33, no.~4, pp. 899--925, 2021.

\bibitem{lian2023learnable}
S.~Lian, J.~Shen, Q.~Liu, Z.~Wang, R.~Yan, and H.~Tang, ``Learnable surrogate gradient for direct training spiking neural networks.'' in \emph{Proceedings of International Joint Conference on Artificial Intelligence (IJCAI)}, 2023, pp. 3002--3010.

\bibitem{liu2018progressive}
C.~Liu, B.~Zoph, M.~Neumann, J.~Shlens, W.~Hua, L.-J. Li, L.~Fei-Fei, A.~Yuille, J.~Huang, and K.~Murphy, ``Progressive neural architecture search,'' in \emph{Proceedings of the European Conference on Computer Vision (ECCV)}, 2018, pp. 19--34.

\bibitem{real2019regularized}
E.~Real, A.~Aggarwal, Y.~Huang, and Q.~V. Le, ``Regularized evolution for image classifier architecture search,'' in \emph{Proceedings of Association for the Advancement of Artificial Intelligence (AAAI)}, vol.~33, no.~01, 2019, pp. 4780--4789.

\bibitem{chen2019detnas}
Y.~Chen, T.~Yang, X.~Zhang, G.~Meng, X.~Xiao, and J.~Sun, ``Detnas: Backbone search for object detection,'' \emph{Proceedings of Advances in Neural Information Processing Systems (NeurIPS)}, vol.~32, 2019.

\bibitem{guo2020hit}
J.~Guo, K.~Han, Y.~Wang, C.~Zhang, Z.~Yang, H.~Wu, X.~Chen, and C.~Xu, ``Hit-detector: Hierarchical trinity architecture search for object detection,'' in \emph{Proceedings of the IEEE/CVF Conference on Computer Vision and Pattern Recognition (CVPR)}, 2020, pp. 11\,405--11\,414.

\bibitem{nekrasov2019fast}
V.~Nekrasov, H.~Chen, C.~Shen, and I.~Reid, ``Fast neural architecture search of compact semantic segmentation models via auxiliary cells,'' in \emph{Proceedings of the IEEE/CVF Conference on Computer Vision and Pattern Recognition (CVPR)}, 2019, pp. 9126--9135.

\bibitem{lin2020graph}
P.~Lin, P.~Sun, G.~Cheng, S.~Xie, X.~Li, and J.~Shi, ``Graph-guided architecture search for real-time semantic segmentation,'' in \emph{Proceedings of the IEEE/CVF Conference on Computer Vision and Pattern Recognition (CVPR)}, 2020, pp. 4203--4212.

\bibitem{xu2019pc}
Y.~Xu, L.~Xie, X.~Zhang, X.~Chen, G.-J. Qi, Q.~Tian, and H.~Xiong, ``Pc-darts: Partial channel connections for memory-efficient architecture search,'' in \emph{Proceedings of International Conference on Learning Representations (ICLR)}, 2019.

\bibitem{chen2019progressive}
X.~Chen, L.~Xie, J.~Wu, and Q.~Tian, ``Progressive differentiable architecture search: Bridging the depth gap between search and evaluation,'' in \emph{Proceedings of the IEEE/CVF International Conference on Computer Vision (ICCV)}, 2019, pp. 1294--1303.

\bibitem{chu2020darts}
X.~Chu, X.~Wang, B.~Zhang, S.~Lu, X.~Wei, and J.~Yan, ``Darts-: Robustly stepping out of performance collapse without indicators,'' in \emph{Proceedings of International Conference on Learning Representations (ICLR)}, 2020.

\bibitem{cheng2020hierarchical}
X.~Cheng, Y.~Zhong, M.~Harandi, Y.~Dai, X.~Chang, H.~Li, T.~Drummond, and Z.~Ge, ``Hierarchical neural architecture search for deep stereo matching,'' \emph{Proceedings of Advances in Neural Information Processing Systems (NeurIPS)}, vol.~33, pp. 22\,158--22\,169, 2020.

\bibitem{shen2024evolutionary}
S.~Shen, R.~Zhang, C.~Wang, R.~Huang, A.~Tuerhong, Q.~Guo, Z.~Lu, J.~Zhang, and L.~Leng, ``Evolutionary spiking neural networks: a survey,'' \emph{Journal of Membrane Computing}, pp. 1--12, 2024.

\bibitem{na2022autosnn}
B.~Na, J.~Mok, S.~Park, D.~Lee, H.~Choe, and S.~Yoon, ``Autosnn: Towards energy-efficient spiking neural networks,'' in \emph{Proceedings of International Conference on Machine Learning (ICML)}.\hskip 1em plus 0.5em minus 0.4em\relax PMLR, 2022, pp. 16\,253--16\,269.

\bibitem{kim2022neural}
Y.~Kim, Y.~Li, H.~Park, Y.~Venkatesha, and P.~Panda, ``Neural architecture search for spiking neural networks,'' in \emph{Proceedings of the European Conference on Computer Vision (ECCV)}.\hskip 1em plus 0.5em minus 0.4em\relax Springer, 2022, pp. 36--56.

\bibitem{wang2023evolving}
G.~Wang, Y.~Sun, S.~Cheng, and S.~Song, ``Evolving connectivity for recurrent spiking neural networks,'' \emph{Proceedings of Advances in Neural Information Processing Systems (NeurIPS)}, vol.~36, pp. 2991--3007, 2023.

\bibitem{gaier2019weight}
A.~Gaier and D.~Ha, ``Weight agnostic neural networks,'' \emph{Proceedings of Advances in Neural Information Processing Systems (NeurIPS)}, vol.~32, 2019.

\bibitem{shen2023brain}
G.~Shen, D.~Zhao, Y.~Dong, and Y.~Zeng, ``Brain-inspired neural circuit evolution for spiking neural networks,'' \emph{Proceedings of the National Academy of Sciences}, vol. 120, no.~39, p. e2218173120, 2023.

\bibitem{xie2023efficient}
Z.~Xie, Z.~Liu, P.~Chen, and J.~Zhang, ``Efficient spiking neural architecture search with mixed neuron models and variable thresholds,'' in \emph{International Conference on Neural Information Processing (ICONIP)}.\hskip 1em plus 0.5em minus 0.4em\relax Springer, 2023, pp. 466--481.

\bibitem{yan2024sampling}
S.~Yan, Q.~Meng, M.~Xiao, Y.~Wang, and Z.~Lin, ``Sampling complex topology structures for spiking neural networks,'' \emph{Neural Networks}, vol. 172, p. 106121, 2024.

\bibitem{yan2024efficient}
J.~Yan, Q.~Liu, M.~Zhang, L.~Feng, D.~Ma, H.~Li, and G.~Pan, ``Efficient spiking neural network design via neural architecture search,'' \emph{Neural Networks}, vol. 173, p. 106172, 2024.

\bibitem{zimmer2019technical}
R.~Zimmer, T.~Pellegrini, S.~F. Singh, and T.~Masquelier, ``Technical report: supervised training of convolutional spiking neural networks with pytorch,'' \emph{arXiv preprint arXiv:1911.10124}, 2019.

\bibitem{wu2021liaf}
Z.~Wu, H.~Zhang, Y.~Lin, G.~Li, M.~Wang, and Y.~Tang, ``Liaf-net: Leaky integrate and analog fire network for lightweight and efficient spatiotemporal information processing,'' \emph{IEEE Transactions on Neural Networks and Learning Systems}, vol.~33, no.~11, pp. 6249--6262, 2021.

\bibitem{fang2021incorporating}
W.~Fang, Z.~Yu, Y.~Chen, T.~Masquelier, T.~Huang, and Y.~Tian, ``Incorporating learnable membrane time constant to enhance learning of spiking neural networks,'' in \emph{Proceedings of the IEEE/CVF International Conference on Computer Vision (ICCV)}, 2021, pp. 2661--2671.

\bibitem{yang2019dashnet}
Z.~Yang, Y.~Wu, G.~Wang, Y.~Yang, G.~Li, L.~Deng, J.~Zhu, and L.~Shi, ``Dashnet: A hybrid artificial and spiking neural network for high-speed object tracking,'' \emph{arXiv preprint arXiv:1909.12942}, 2019.

\bibitem{lee2020spike}
C.~Lee, A.~K. Kosta, A.~Z. Zhu, K.~Chaney, K.~Daniilidis, and K.~Roy, ``Spike-flownet: event-based optical flow estimation with energy-efficient hybrid neural networks,'' in \emph{Proceedings of the European Conference on Computer Vision (ECCV)}.\hskip 1em plus 0.5em minus 0.4em\relax Springer, 2020, pp. 366--382.

\bibitem{10347028}
X.~Chen, Q.~Yang, J.~Wu, H.~Li, and K.~C. Tan, ``A hybrid neural coding approach for pattern recognition with spiking neural networks,'' \emph{IEEE Transactions on Pattern Analysis and Machine Intelligence}, vol.~46, no.~5, pp. 3064--3078, 2024.

\bibitem{gallego2020event}
G.~Gallego, T.~Delbr{\"u}ck, G.~Orchard, C.~Bartolozzi, B.~Taba, A.~Censi, S.~Leutenegger, A.~J. Davison, J.~Conradt, K.~Daniilidis \emph{et~al.}, ``Event-based vision: A survey,'' \emph{IEEE Transactions on Pattern Analysis and Machine Intelligence}, vol.~44, no.~1, pp. 154--180, 2020.

\bibitem{merolla2014million}
P.~A. Merolla, J.~V. Arthur, R.~Alvarez-Icaza, A.~S. Cassidy, J.~Sawada, F.~Akopyan, B.~L. Jackson, N.~Imam, C.~Guo, Y.~Nakamura \emph{et~al.}, ``A million spiking-neuron integrated circuit with a scalable communication network and interface,'' \emph{Science}, vol. 345, no. 6197, pp. 668--673, 2014.

\bibitem{roy2019towards}
K.~Roy, A.~Jaiswal, and P.~Panda, ``Towards spike-based machine intelligence with neuromorphic computing,'' \emph{Nature}, vol. 575, no. 7784, pp. 607--617, 2019.

\bibitem{9695196}
L.~Zhu, S.~Dong, J.~Li, T.~Huang, and Y.~Tian, ``Ultra-high temporal resolution visual reconstruction from a fovea-like spike camera via spiking neuron model,'' \emph{IEEE Transactions on Pattern Analysis and Machine Intelligence}, vol.~45, no.~1, pp. 1233--1249, 2023.

\bibitem{9767613}
R.~W. Baldwin, R.~Liu, M.~Almatrafi, V.~Asari, and K.~Hirakawa, ``Time-ordered recent event (tore) volumes for event cameras,'' \emph{IEEE Transactions on Pattern Analysis and Machine Intelligence}, vol.~45, no.~2, pp. 2519--2532, 2023.

\bibitem{10019594}
Y.~Zheng, L.~Zheng, Z.~Yu, T.~Huang, and S.~Wang, ``Capture the moment: High-speed imaging with spiking cameras through short-term plasticity,'' \emph{IEEE Transactions on Pattern Analysis and Machine Intelligence}, vol.~45, no.~7, pp. 8127--8142, 2023.

\bibitem{kugele2020efficient}
A.~Kugele, T.~Pfeil, M.~Pfeiffer, and E.~Chicca, ``Efficient processing of spatio-temporal data streams with spiking neural networks,'' \emph{Frontiers in Neuroscience}, vol.~14, p. 439, 2020.

\bibitem{li2024efficient}
B.~Li, L.~Leng, S.~Shen, K.~Zhang, J.~Zhang, J.~Liao, and R.~Cheng, ``Efficient deep spiking multilayer perceptrons with multiplication-free inference,'' \emph{IEEE Transactions on Neural Networks and Learning Systems}, 2024.

\bibitem{kim2020spiking}
S.~Kim, S.~Park, B.~Na, and S.~Yoon, ``Spiking-yolo: Spiking neural network for energy-efficient object detection,'' in \emph{Proceedings of Association for the Advancement of Artificial Intelligence (AAAI)}, vol.~34, no.~07, 2020, pp. 11\,270--11\,277.

\bibitem{zhang2024automotive}
H.~Zhang, Y.~Li, L.~Leng, K.~Che, Q.~Liu, Q.~Guo, J.~Liao, and R.~Cheng, ``Automotive object detection via learning sparse events by spiking neurons,'' \emph{IEEE Transactions on Cognitive and Developmental Systems}, 2024.

\bibitem{zhang2024accurate}
R.~Zhang, L.~Leng, K.~Che, H.~Zhang, J.~Cheng, Q.~Guo, J.~Liao, and R.~Cheng, ``Accurate and efficient event-based semantic segmentation using adaptive spiking encoder--decoder network,'' \emph{IEEE Transactions on Neural Networks and Learning Systems}, 2024.

\bibitem{8660483}
F.~Paredes-Vallés, K.~Y.~W. Scheper, and G.~C. H.~E. de~Croon, ``Unsupervised learning of a hierarchical spiking neural network for optical flow estimation: From events to global motion perception,'' \emph{IEEE Transactions on Pattern Analysis and Machine Intelligence}, vol.~42, no.~8, pp. 2051--2064, 2020.

\bibitem{10130595}
S.~Liu and P.~L. Dragotti, ``Sensing diversity and sparsity models for event generation and video reconstruction from events,'' \emph{IEEE Transactions on Pattern Analysis and Machine Intelligence}, vol.~45, no.~10, pp. 12\,444--12\,458, 2023.

\bibitem{zhu2018realtime}
A.~Z. Zhu, Y.~Chen, and K.~Daniilidis, ``Realtime time synchronized event-based stereo,'' in \emph{Proceedings of the European Conference on Computer Vision (ECCV)}, 2018, pp. 433--447.

\bibitem{zhou2018semi}
Y.~Zhou, G.~Gallego, H.~Rebecq, L.~Kneip, H.~Li, and D.~Scaramuzza, ``Semi-dense 3d reconstruction with a stereo event camera,'' in \emph{Proceedings of the European Conference on Computer Vision (ECCV)}, 2018, pp. 235--251.

\bibitem{tulyakov2019learning}
S.~Tulyakov, F.~Fleuret, M.~Kiefel, P.~Gehler, and M.~Hirsch, ``Learning an event sequence embedding for dense event-based deep stereo,'' in \emph{Proceedings of the IEEE/CVF International Conference on Computer Vision (ICCV)}, 2019, pp. 1527--1537.

\bibitem{ahmed2021deep}
S.~H. Ahmed, H.~W. Jang, S.~N. Uddin, and Y.~J. Jung, ``Deep event stereo leveraged by event-to-image translation,'' in \emph{Proceedings of Association for the Advancement of Artificial Intelligence (AAAI)}, vol.~35, no.~2, 2021, pp. 882--890.

\bibitem{mostafavi2021event}
M.~Mostafavi, K.-J. Yoon, and J.~Choi, ``Event-intensity stereo: Estimating depth by the best of both worlds,'' in \emph{Proceedings of the IEEE/CVF International Conference on Computer Vision (ICCV)}, 2021, pp. 4258--4267.

\bibitem{zhang2022discrete}
K.~Zhang, K.~Che, J.~Zhang, J.~Cheng, Z.~Zhang, Q.~Guo, and L.~Leng, ``Discrete time convolution for fast event-based stereo,'' in \emph{Proceedings of the IEEE/CVF Conference on Computer Vision and Pattern Recognition (CVPR)}, 2022, pp. 8676--8686.

\bibitem{ranccon2022stereospike}
U.~Ran{\c{c}}on, J.~Cuadrado-Anibarro, B.~R. Cottereau, and T.~Masquelier, ``Stereospike: Depth learning with a spiking neural network,'' \emph{IEEE Access}, vol.~10, pp. 127\,428--127\,439, 2022.

\bibitem{bellec2018long}
G.~Bellec, D.~Salaj, A.~Subramoney, R.~Legenstein, and W.~Maass, ``Long short-term memory and learning-to-learn in networks of spiking neurons,'' \emph{Proceedings of Advances in Neural Information Processing Systems (NeurIPS)}, vol.~31, 2018.

\bibitem{zenke2018superspike}
F.~Zenke and S.~Ganguli, ``Superspike: Supervised learning in multilayer spiking neural networks,'' \emph{Neural Computation}, vol.~30, no.~6, pp. 1514--1541, 2018.

\bibitem{ioffe2015batch}
S.~Ioffe and C.~Szegedy, ``Batch normalization: Accelerating deep network training by reducing internal covariate shift,'' in \emph{Proceedings of International Conference on Machine Learning (ICML)}.\hskip 1em plus 0.5em minus 0.4em\relax PMLR, 2015, pp. 448--456.

\bibitem{2019Towards}
J.~Pei, L.~Deng, S.~Song, M.~Zhao, and L.~Shi, ``Towards artificial general intelligence with hybrid tianjic chip architecture,'' \emph{Nature}, vol. 572, no. 7767, p. 106, 2019.

\bibitem{horowitz20141}
M.~Horowitz, ``1.1 computing's energy problem (and what we can do about it),'' in \emph{IEEE International Solid-state Circuits Conference Digest of Technical Papers (ISSCC)}.\hskip 1em plus 0.5em minus 0.4em\relax IEEE, 2014, pp. 10--14.

\bibitem{tulyakov2018practical}
S.~Tulyakov, A.~Ivanov, and F.~Fleuret, ``Practical deep stereo (pds): Toward applications-friendly deep stereo matching,'' \emph{Proceedings of Advances in Neural Information Processing Systems (NeurIPS)}, vol.~31, 2018.

\bibitem{wang2019event}
L.~Wang, Y.-S. Ho, K.-J. Yoon \emph{et~al.}, ``Event-based high dynamic range image and very high frame rate video generation using conditional generative adversarial networks,'' in \emph{Proceedings of the IEEE/CVF Conference on Computer Vision and Pattern Recognition (CVPR)}, 2019, pp. 10\,081--10\,090.

\bibitem{cifar}
\BIBentryALTinterwordspacing
A.~Krizhevsky, V.~Nair, and G.~Hinton, ``Cifar-10 (canadian institute for advanced research).'' [Online]. Available: \url{http://www.cs.toronto.edu/~kriz/cifar.html}
\BIBentrySTDinterwordspacing

\bibitem{deng2009imagenet}
J.~Deng, W.~Dong, R.~Socher, L.-J. Li, K.~Li, and L.~Fei-Fei, ``Imagenet: A large-scale hierarchical image database,'' in \emph{Proceedings of the IEEE/CVF Conference on Computer Vision and Pattern Recognition (CVPR)}.\hskip 1em plus 0.5em minus 0.4em\relax Ieee, 2009, pp. 248--255.

\bibitem{wu2021tandem}
J.~Wu, Y.~Chua, M.~Zhang, G.~Li, H.~Li, and K.~C. Tan, ``A tandem learning rule for effective training and rapid inference of deep spiking neural networks,'' \emph{IEEE Transactions on Neural Networks and Learning Systems}, vol.~34, no.~1, pp. 446--460, 2021.

\bibitem{ptflops}
\BIBentryALTinterwordspacing
V.~Sovrasov. (2019) Flops counter for convolutional networks in pytorch framework. [Online]. Available: \url{https://github.com/sovrasov/flops-counter.pytorch/}
\BIBentrySTDinterwordspacing

\end{thebibliography}



%




\begin{IEEEbiography}[{\includegraphics[width=1in,height=1.25in,clip,keepaspectratio]{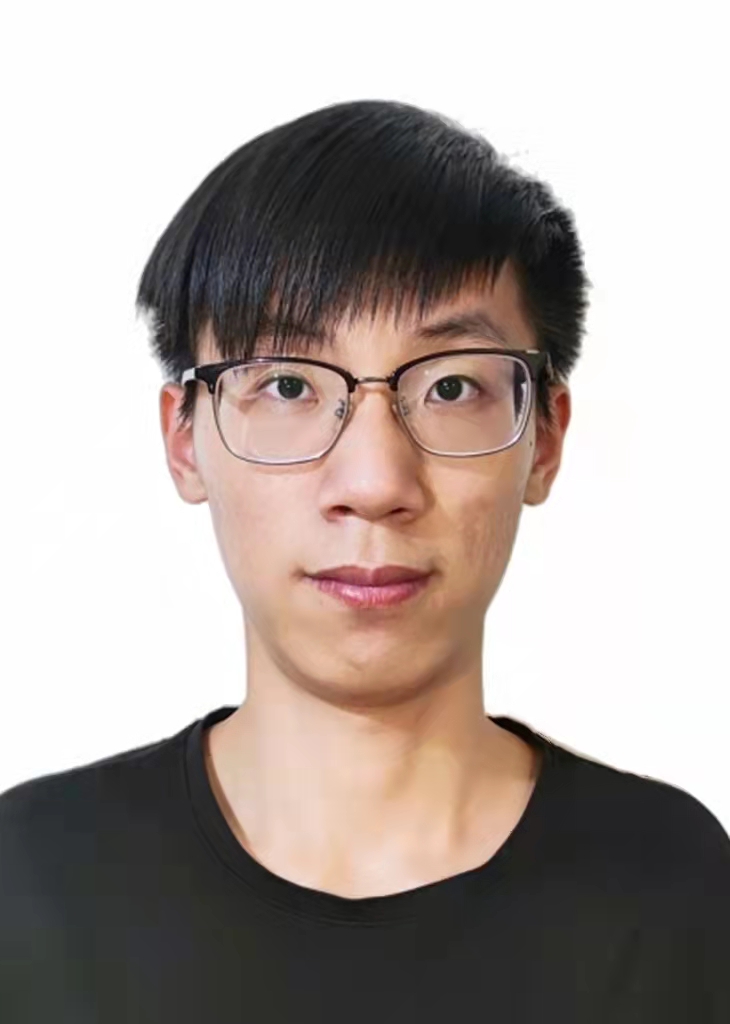}}]
{Kaiwei Che} received the BEng degree from Shenzhen University, in 2020, and the MEng degree from the Southern University of Science and Technology, in 2023. He is currently pursuing Ph.D. degree at School of Computer Science, Peking University. His research interests include deep learning and brain-inspired algorithms.
\end{IEEEbiography}
\begin{IEEEbiography}[{\includegraphics[width=1in,height=1.25in,clip,keepaspectratio]{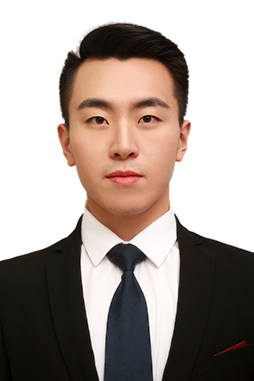}}]
{Zhaokun Zhou} received the BE and ME degrees from Chongqing University, Chongqing, China, in 2018 and 2021. He is currently pursuing Ph.D. degree at School of Computer Science, Peking University. His research interests encompass spiking deep learning, neuromorphic computing, and computer vision.
\end{IEEEbiography}

\begin{IEEEbiography}[{\includegraphics[width=1in,height=1.25in,clip,keepaspectratio]{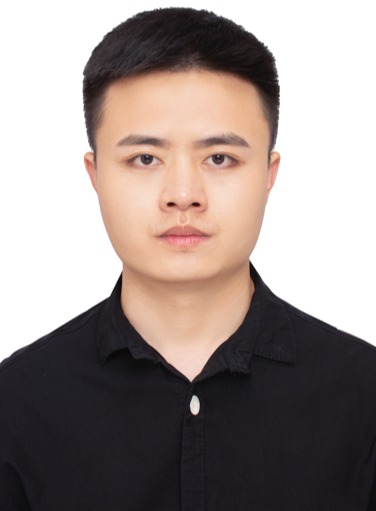}}]
{Li Yuan} received the BEng degree from the University of Science and Technology of China, in 2017, and the PhD degree from the National University of Singapore, in 2021. He is currently a tenure-track Assistant Professor with the School of Electrical and Computer Engineering, Peking University. He has published more than 20 papers on top conferences/journals. His research interests include deep learning, image processing,
and computer vision.
\end{IEEEbiography}

\begin{IEEEbiography}[{\includegraphics[width=1in,height=1.25in,clip,keepaspectratio]{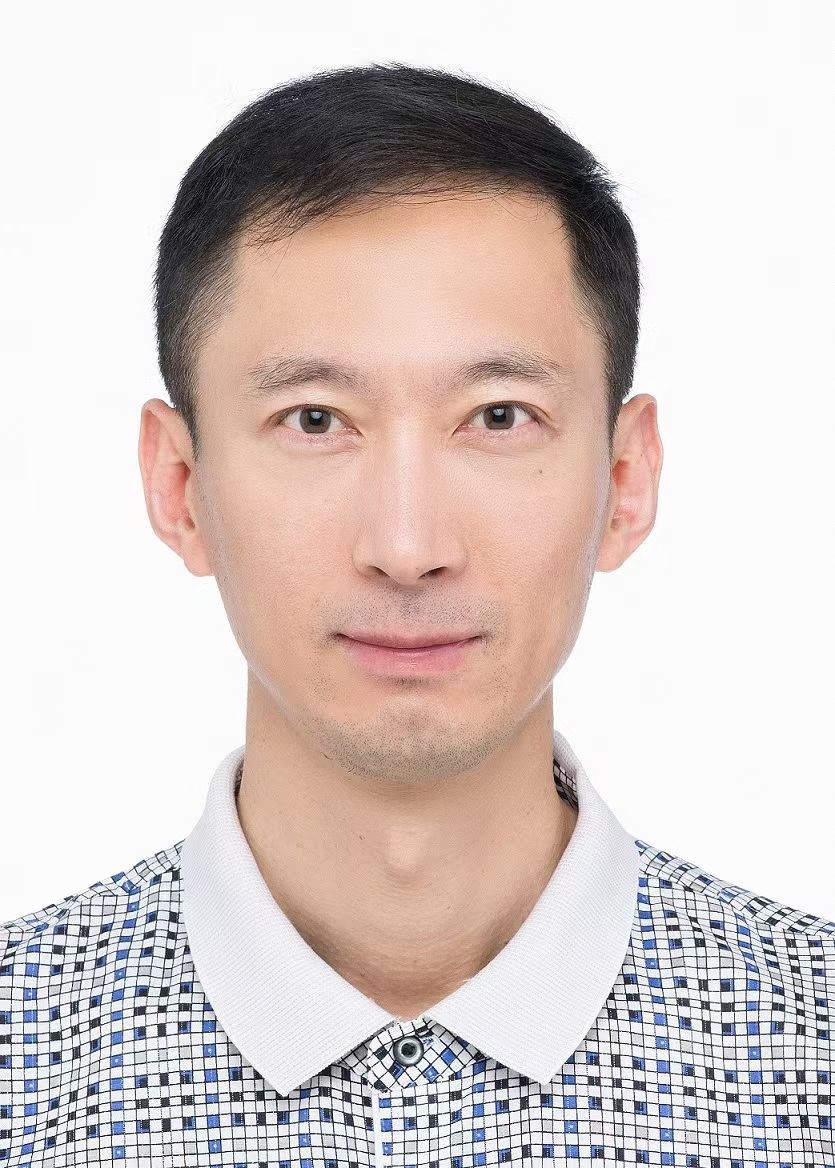}}]
{Jianguo Zhang} is currently a Professor in Department of Computer Science and Engineering, Southern University of Science and Technology. Previously, he was a Reader in Computing, School of Science and Engineering, University of Dundee, UK. He received a PhD in National Lab of Pattern Recognition, Institute of Automation, Chinese Academy of Sciences, Beijing, China, 2002. His research interests include object recognition, medical image analysis, machine learning and computer vision. He is a senior member of the IEEE and serves as an Associate Editor of IEEE Trans on Multimedia
and computer vision.
\end{IEEEbiography}

\begin{IEEEbiography}[{\includegraphics[width=1in,height=1.25in,clip,keepaspectratio]{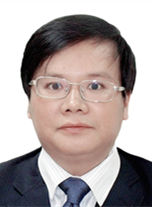}}]
{Yonghong Tian} (S’00-M’06-SM’10-F’22) is currently the Dean of School of Electronics and Computer Engineering, a Boya Distinguished Professor with the School of Computer Science, Peking University, China, and is also the deputy director of Artificial Intelligence Research, PengCheng Laboratory, Shenzhen, China. His research interests include neuromorphic vision, distributed machine learning and multimedia big data. He is the author or coauthor of over 300 technical articles in refereed journals and conferences. Prof. Tian was/is an Associate Editor of IEEE TCSVT (2018.1-2021.12), IEEE TMM (2014.8-2018.8), IEEE Multimedia Mag. (2018.1-2022.8), and IEEE Access (2017.1-2021.12). He co-initiated IEEE Int’l Conf. on Multimedia Big Data (BigMM) and served as the TPC Co-chair of BigMM 2015, and aslo served as the Technical Program Cochair of IEEE ICME 2015, IEEE ISM 2015 and IEEE MIPR 2018/2019, and General Co-chair of IEEE MIPR 2020 and ICME2021. He is a TPC Member of more than ten conferences such as CVPR, ICCV, ACM KDD, AAAI, ACM MM and ECCV. He was the recipient of the Chinese National Science Foundation for Distinguished Young Scholars in 2018, two National Science and Technology Awards and three ministerial-level awards in China, and obtained the 2015 EURASIP Best Paper Award for Journal on Image and Video Processing, and the best paper award of IEEE BigMM 2018, and the 2022 IEEE SA Standards Medallion and SA Emerging Technology Award. He is a Fellow of IEEE, a senior member of CIE and CCF, a member of ACM.
\end{IEEEbiography}

\begin{IEEEbiography}[{\includegraphics[width=1in,height=1.25in,clip,keepaspectratio]{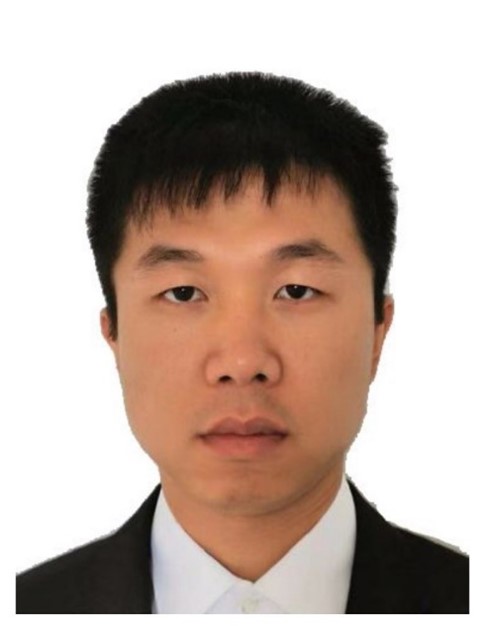}}]
{Luziwei Leng} is currently a Principal Engineer at the Advanced Computing and Storage Lab of Huawei Technologies Co., Ltd and an industry supervisor of the joint-master program between Huawei and the Southern University of Science and Technology, Shenzhen, China. He received a Doctoral degree from the Department of Physics, Heidelberg University, Germany in 2019. His research focuses on brain-inspired computing and machine learning, including spiking neural networks, event-based vision, bio-inspired learning rules and evolutionary algorithms, etc. He is an editorial board member of the Frontiers in Neuroscience (Neuromorphic Engineering).
\end{IEEEbiography}




\end{document}